\documentclass[]{mosi}

\usepackage{helvet}

\usepackage{amsmath} 
\usepackage{natbib}
\usepackage{graphicx}
\usepackage{subcaption} 

\usepackage[toc,page,header]{appendix}
\usepackage[utf8]{inputenc} 
\usepackage[T1]{fontenc}    
\usepackage{hyperref}       
\usepackage{url}            
\usepackage{booktabs}       
\usepackage{lmodern}        
\usepackage{amsfonts}       
\usepackage{nicefrac}       
\usepackage{microtype}      
\usepackage{wrapfig}

\usepackage{amssymb}        
\usepackage{titletoc}
\usepackage{minitoc}

\usepackage{array}
\usepackage{etoolbox}

\definecolor{lightblue}{RGB}{200, 230, 255}  
\definecolor{headerblue}{RGB}{150, 200, 255} 

\usepackage{pgfplots}
\usepackage{xcolor}
\usepackage{float}
\usepackage{comment}
\usepackage{multirow}
\usepackage{makecell}
\usepackage{siunitx}
\usepackage{tikz}
\usepackage{pgf-pie}
\usepackage[export]{adjustbox}

\usepackage{ragged2e}
\usepackage{tabularx}
\usepackage{caption}
\usepackage{enumitem}
\usepackage{pifont}
\usepackage[hang,flushmargin]{footmisc}

\usepackage{tcolorbox}
\tcbuselibrary{breakable}
\tcbuselibrary{skins}

\usepackage{tabularx}
\usepackage{listings}

\usepackage{threeparttable}
\usepackage{setspace}
\usepackage[scheme=plain]{ctex} 

\definecolor{MossCyan}{HTML}{82D9FF} 
\definecolor{MossBlue}{HTML}{82B1FF}

\definecolor{ForestGreen}{RGB}{34, 139, 34}
\definecolor{Red}{RGB}{255, 0, 0}

\definecolor{tickG}{rgb}{0.1, 0.588, 0.1}
\definecolor{crossR}{rgb}{0.588, 0.1, 0.1}
\newcommand{\cmark}{\textcolor{tickG}{\ding{52}}}
\newcommand{\xmark}{\textcolor{crossR}{\ding{56}}}

\definecolor{frenchblue}{rgb}{0.0, 0.45, 0.73}
\definecolor{babyblue}{rgb}{0.54, 0.81, 0.94}
\definecolor{classicrose}{rgb}{0.98, 0.8, 0.91}
\definecolor{beige}{rgb}{0.96, 0.96, 0.86}
\definecolor{forestgreen}{HTML}{2e7d43}

\definecolor{blue1}{HTML}{91BBE6}
\definecolor{blue2}{HTML}{3F90E0}
\definecolor{blue3}{HTML}{316FAD}

\definecolor{color1}{HTML}{FF9999}
\definecolor{color2}{HTML}{FF6666}
\definecolor{color3}{HTML}{FF3333}
\definecolor{color4}{HTML}{E60000}
\definecolor{color5}{HTML}{B30000}
\definecolor{color6}{HTML}{8CD98C}
\definecolor{color7}{HTML}{53c653}
\definecolor{color8}{HTML}{00B050}
\definecolor{color9}{HTML}{2d862d}
\definecolor{color10}{HTML}{206020}
\definecolor{color11}{HTML}{cca300}

\newtcolorbox{promptbox}[2][]{
    colback=white,
    coltext=black,
    arc=3mm,
    boxrule=0.5pt,
    colframe=black!60!white,
    title={#2},
    colbacktitle=black,
    coltitle=white,
    fonttitle=\bfseries,
    top=8pt,
    bottom=8pt,
    left=10pt,
    right=10pt,
    breakable,
    before upper={%
        \linespread{1}\selectfont
        \setlength{\parskip}{1ex plus 0.2ex minus 0.2ex}%
        \setlength{\parindent}{0pt}%
    },
    #1
}

\lstdefinelanguage{json}{
    basicstyle=\ttfamily\scriptsize, 
    numbers=none,
    stepnumber=1,
    numbersep=5pt,
    backgroundcolor=\color{white},   
    showstringspaces=false,
    breaklines=true,                 
    frame=single,                    
    framexleftmargin=5pt,
    keepspaces=true,
    commentstyle=\color{gray},       
    keywordstyle=\color{blue},       
    stringstyle=\color{violet},      
    morestring=[b]",
    morestring=[d]',
    escapeinside={(*@}{@*)}
}

\title{AgentLongBench: A Controllable Long Benchmark For Long-Contexts Agents via Environment Rollouts}

\author{
Shicheng Fang$^{1,2}$ \hspace{.3em}
Yuxin Wang$^{1}$\hspace{.3em}
Xiaoran Liu$^{1,2}$ \hspace{.1em}
Jiahao Lu$^{1,2}$
\\
\textbf{
Chuanyuan Tan$^{3}$ \hspace{.1em}
Xinchi Chen$^{1}$ \hspace{.1em}
Yining Zheng$^{1,\dagger}$ \hspace{.1em}
Xuanjing Huang$^{1}$ \hspace{.1em}
}
\\
\textbf{
Xipeng Qiu$^{1,2,\dagger}$ \hspace{.2em}
}
\\
[1ex]
$^{1}$Fudan University   
$^{2}$Shanghai Innovation Institute   
$^{3}$Soochow University
\\
}

\abstract{
\begin{abstract}
'The evolution of Large Language Models (LLMs) into autonomous agents necessitates the management of extensive, dynamic contexts. Current benchmarks, however, remain largely static, relying on passive retrieval tasks that fail to simulate the complexities of agent-environment interaction, such as non-linear reasoning and iterative feedback. To address this, we introduce \textbf{AgentLongBench}, which evaluates agents through simulated environment rollouts based on Lateral Thinking Puzzles. This framework generates rigorous interaction trajectories across knowledge-intensive and knowledge-free scenarios. Experiments with state-of-the-art models and memory systems (32K to 4M tokens) expose a critical weakness: while adept at static retrieval, agents struggle with the dynamic information synthesis essential for workflows. Our analysis indicates that this degradation is driven by the minimum number of tokens required to resolve a query. This factor explains why the high information density inherent in massive tool responses poses a significantly greater challenge than the memory fragmentation typical of long-turn dialogues.
\end{abstract}
}

\checkdata[Correspondence]{\url{scfang25@m.fudan.edu.cn}, \url{ynzheng@fudan.edu.cn}, \url{xpqiu@fudan.edu.cn}}
\checkdata[Repository]{\url{https://github.com/euReKa025/AgentLongBench}}
\checkdata[Benchmark]{\url{https://huggingface.co/datasets/ign1s/AgentLongBench}}

\begin{document}
\maketitle
\begingroup
\renewcommand{\thefootnote}{\fnsymbol{footnote}}
\endgroup



\section{Introduction}

\begin{figure}[t]
    \centering
    \includegraphics[width=0.48\linewidth]{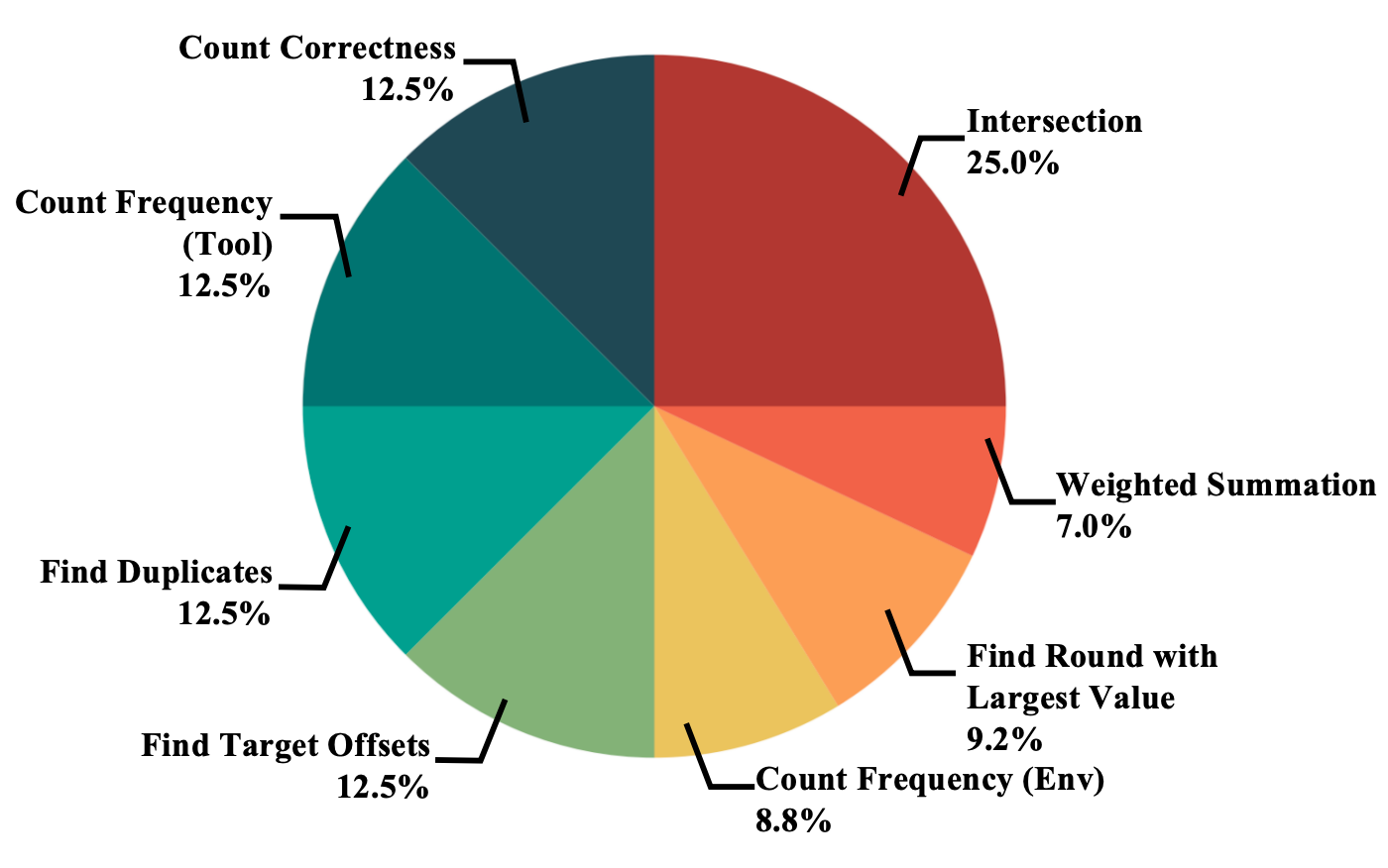}
    \caption{The distribution of distinct question types under the \textit{Knowledge-Intensive} setting with \textit{Concise-Response} formatting in our dataset.}
    \label{fig:distributionofsamples}
\end{figure}

The rapid evolution of Large Language Models (LLMs) from static chatbots to autonomous agents demands capabilities that extend far beyond simple conversation. Agents must synthesize vast amounts of historical information to execute complex workflows; failure to do so leads to fragmented reasoning, hallucinations, and planning errors. Although the community has actively expanded context windows and developed retrieval-augmented architectures, evaluation frameworks remain disproportionately focused on passive reading comprehension.

Prevalent benchmarks still prioritize static retrieval, often asking models to locate isolated facts within artificially concatenated documents. This paradigm misses the core of agentic behavior. Real-world problem solving involves dynamic tool usage and non-linear reasoning, creating a context that evolves based on the agent's own decisions. Unlike the static User-AI dialogues found in current datasets, autonomous operations generate unique AI-Environment trajectories. These scenarios demand active state tracking through iterative feedback loops, a challenge that fixed, human-authored texts fail to capture.

\begin{table*}[htbp]
    \centering
    \resizebox{\textwidth}{!}{\begin{tabular}{l|c|c|c|c|c|c|c|}
    \toprule
         & Max Length & Evolving Context&Knowledge-Free&Det. Ground Truth&State Tracking&Tool Use
Logs&  Controllability\\
         \midrule
         NeedleBench\citep{li2024needlebench}  &  128K&  \xmark&\cmark&\cmark&\xmark&\xmark& \xmark\\
         BABILong\citep{NEURIPS2024_babilong}  &  10M&  \xmark&\xmark&\cmark&\xmark&\xmark& \xmark\\
         LV-Eval\citep{yuan2024lv}  &  256K&  \xmark&\xmark&\xmark&\xmark&\xmark& \xmark\\
         RULER\cite{hsiehruler}  &  128K&  \xmark&\cmark&\cmark&\cmark&\xmark& \cmark\\
         L-Eval\citep{an2024eval}  &  200K&  \xmark&\xmark&\xmark&\xmark&\xmark& \xmark\\
      Michelangelo\citep{Vodrahalli2024MichelangeloLC}  &  128K&  \cmark&\cmark&\cmark&\cmark&\xmark& \cmark\\
       Loong\citep{wang2024leave}  &  250K&  \xmark&\xmark&\xmark&\xmark&\xmark& \xmark\\
       BAMBOO\citep{dong2024bamboo}  & 16K  &  \xmark&\xmark&\cmark&\xmark&\xmark& \xmark\\
       $\infty$-Bench\citep{zhang2024bench}  & 200K  &  \xmark&\xmark&\cmark&\cmark&\xmark& \xmark\\
       LooGLE\citep{li2024loogle}  &  80K&  \xmark&\xmark&\xmark&\xmark&\xmark& \xmark\\
       NoCha\citep{Karpinska2024OneTA}  &  336K&  \xmark&\xmark&\cmark&\xmark&\xmark& \xmark\\
      LongBenchV2\citep{bai2024longbench2}  &  128K&  \xmark&\xmark&\cmark&\xmark&\xmark& \xmark\\
        MRCR(OpenAI)\citep{openai_mrcr}  &  1M&  \xmark&\xmark&\cmark&\xmark&\xmark& \xmark\\
       NoLiMa\citep{modarressi2025nolima}  &  32K&  \xmark&\xmark&\cmark&\xmark&\xmark& \xmark\\
        HELMET\citep{yen2025helmet}  &  128K&  \xmark&\xmark&\xmark&\xmark&\xmark& \xmark\\
         LIFBENCH\citep{wu2025lifbench}  &  128K&  \xmark&\cmark&\cmark&\xmark&\xmark& \xmark\\
         LongBioBench\citep{yang2025controllable}  &  128K&  \xmark&\xmark&\cmark&\xmark&\xmark& \cmark\\
      LooGLE V2\citep{he2025loogle}  &  2M &  \cmark&\xmark&\cmark&\cmark&\xmark& \xmark\\
         \midrule
       LocoMo\citep{maharana2024evaluating}  &  16K&  \cmark&\xmark&\cmark&\cmark&\xmark& \xmark\\
       LongMemEval\citep{wu2024longmemeval}  &  2M&  \xmark&\xmark&\xmark&\cmark&\xmark& \xmark\\
       MemoryAgentBench\citep{hu2025evaluating}  &  2M&  \xmark&\xmark&\cmark&\cmark&\xmark& \xmark\\
       \midrule
       AgentLongBench (ours)  & $\infty$ (4M) & \cmark&\cmark& \cmark&\cmark&\cmark& \cmark\\
       \bottomrule
    \end{tabular}}
    \caption{Comparison of long-context benchmarks. Our published dataset ends at 4M.}
    \label{tab:comparisonswithpreviousworks}
\end{table*}

We introduce \textbf{AgentLongBench} to evaluate agents through rigorous environment rollouts. True agentic capabilities emerge within evolving contexts that mirror the causality of real-world workflows, rather than in static reading comprehension tasks. Consequently, our setting forces agents to parse high-density, machine-generated tool logs instead of natural text distractors, requiring precise state tracking to manage dynamic information updates. To guarantee validity, we rely on deterministic ground truth derived from the environment’s logic and incorporate a Knowledge-Free setting to eliminate parametric memory bias. Furthermore, the framework emphasizes controllability and extensibility \citep{yang2025controllable} to facilitate fine-grained diagnosis of failure modes. The benchmark comprises 8 tasks in each setting. Figure~\ref{fig:distributionofsamples} illustrates the representative task distribution, and we provide full statistics for all settings in Appendix~\ref{sec:appendix_distributions}. Table~\ref{tab:comparisonswithpreviousworks} contrasts our approach with prior benchmarks, highlighting how AgentLongBench uniquely targets the requirements of autonomous operation. A comprehensive review of these related works is available in Appendix~\ref{appendix:relatedworks}.

In summary, the main contributions of our work are as follows:
\begin{itemize}
    \item We introduce \textbf{AgentLongBench}, a controllable benchmark built on simulated environment rollouts that generates dynamic interaction trajectories for evaluating long-horizon consistency and planning.
    \item We propose a comprehensive evaluation taxonomy with \textbf{32 distinct question types}, spanning 2 settings, 2 interaction formats, and 8 tasks, and covering context lengths \textbf{from 32K to 4M tokens}, with 800 samples in each length.
    \item Through extensive evaluations on state-of-the-art LLMs and memory frameworks, we reveal fundamental failure modes in long-horizon agentic reasoning: (i) a strong reliance on parametric knowledge in Knowledge-Free settings; (ii) a trade-off between long-turn interaction trajectories and dense single-turn outputs; and (iii) a substantially higher reasoning burden for high-density tool logs, captured by the notion of a minimum token requirement.
\end{itemize}

\section{AgentLongBench}

\begin{figure*}[t]
    \centering
    \includegraphics[width=0.95\linewidth]{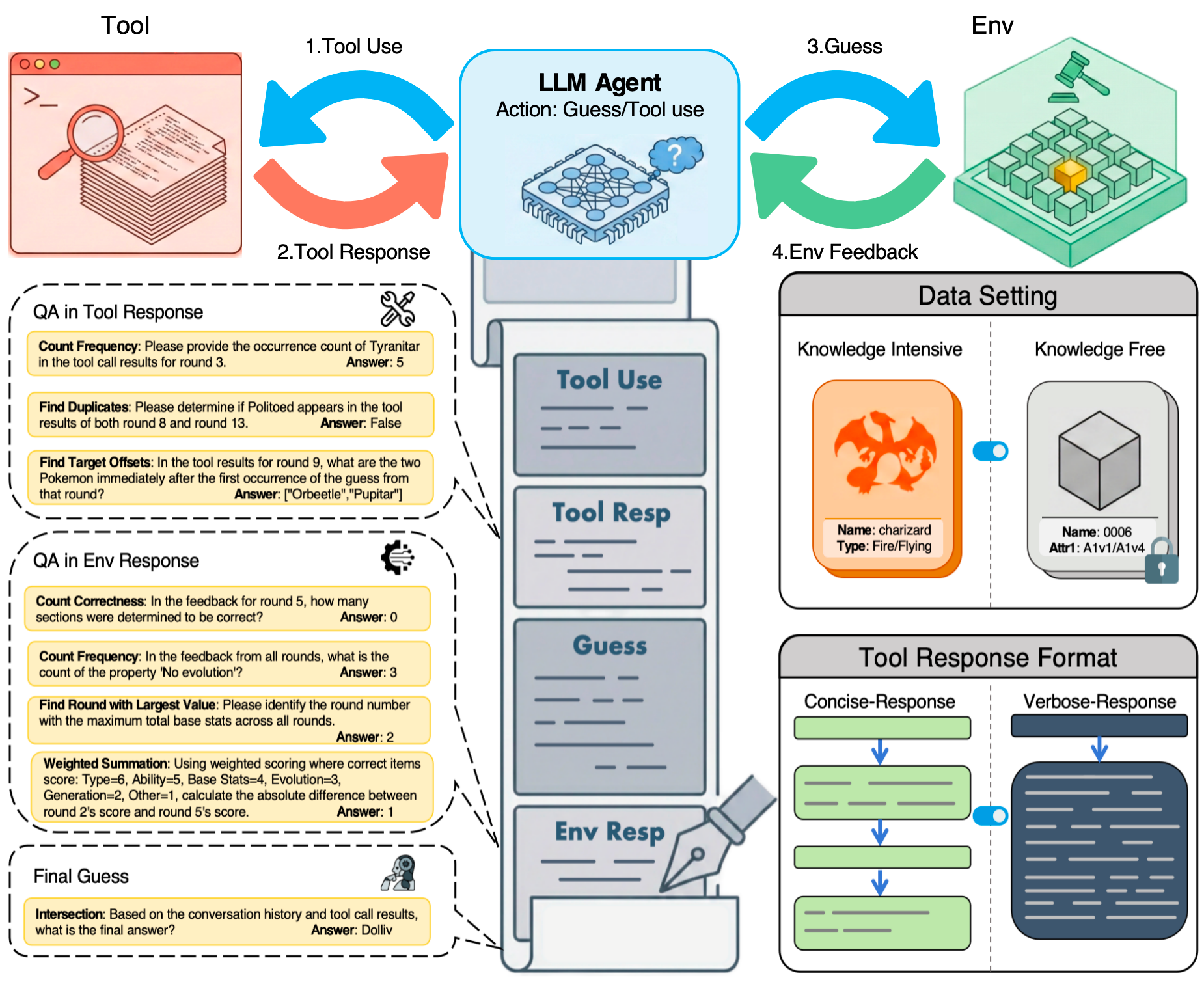}
    \caption{Overview of the Data Construction Pipeline for AgentLongBench. The dataset is constructed by simulating Environment Rollouts (Top), which capture the iterative interaction logs between an LLM agent, tools, and environmental feedback. These trajectories are then used to derive three categories of QA tasks (Left): QA in Tool Response, QA in Environment Response, and Final Guess. The construction process incorporates two data settings (Right Top) to distinguish between \textit{Knowledge-Intensive} and \textit{Knowledge-Free} scenarios, and two tool response formats (Right Bottom)—\textit{Concise} vs. \textit{Verbose}—to simulate different context densities and noise levels.}
    \label{fig:methodology_overview}
\end{figure*}

AgentLongBench shifts the evaluation paradigm from static text processing to dynamic interaction through simulated environment rollouts, with an overview shown in Figure~\ref{fig:methodology_overview}. Instead of relying on concatenated documents, this framework captures the continuous exchange between an agent and a reactive system. By simulating the temporal evolution of context, we preserve the causal dependencies inherent in real-world applications, which are often lost in static reading comprehension tasks.

The foundation of this framework is a Lateral Thinking Puzzle environment \citep{oxfordlateralthinking}, selected for its rigorous demands on iterative inquiry and logical deduction. As the agent hypothesizes and receives feedback, the context expands procedurally based on the environment's deterministic rules. This generation mechanism ensures that the resulting long-context trajectories are verifiable and logically consistent, avoiding the artifacts common in benchmarks constructed via arbitrary document insertion.

We structure the benchmark around two orthogonal dimensions, creating four distinct experimental configurations. To decouple reasoning skills from parametric memory, the first dimension contrasts Knowledge-Intensive with Knowledge-Free scenarios. Simultaneously, we examine the trade-off between temporal span and information density by varying the interaction format between Concise responses and Verbose responses. This cross-dimensional design ensures that performance degradation can be traced directly to either memory retention failures or information overload.

\subsection{Lateral Thinking Puzzle Environment}
The core testbed utilizes Lateral Thinking Puzzles, where agents must reconstruct a hidden state by satisfying a series of logical constraints. This iterative inquiry process contrasts sharply with static QA, serving as a rigorous proxy for the complex, investigative workflows typical of autonomous agents.

\subsubsection{Puzzle Formulation}
We formulate the puzzle as a target identification process under a closed-world assumption. The environment initializes a finite set of items, each defined by a unique vector of attributes ranging from categorical properties (e.g., Type, Abilities) to numerical statistics. One item is designated as the hidden target. In the Knowledge-Intensive setting, we instantiate this item set using the Pokémon\citep{pokemon} dataset (pre-July 2025), ensuring that no two items share identical attribute profiles. The agent operates as an investigator, engaging in a goal-oriented dialogue to bridge the information gap between its current belief state and the ground truth. The puzzle unfolds as the agent continuously queries the environment to isolate the target.

\subsubsection{Environment Response}
The environment functions as a deterministic oracle that holds the ground truth. To simulate rigorous agentic constraints, it parses natural language inquiries and returns precise feedback rather than open-ended narrative descriptions. Responses are typically limited to binary validation (Yes/No) or relational operators for numeric values. Crucially, upon an incorrect guess, the environment generates a comprehensive attribute-wise evaluation. It returns the full profile of the hypothesized item, explicitly annotating the relationship between each attribute and the ground truth—confirming categorical matches while providing directional constraints for numerical discrepancies. This feedback mechanism enforces strict state tracking, as the agent must update its history of valid constraints to iteratively narrow the search space.

\subsubsection{Tool Response}
Beyond direct interaction with the environment, the agent is equipped with auxiliary tools to aid in state refinement. The \textit{Tool Response} represents the output from these external function calls. We provide two search tools corresponding to the two formatting strategies described in Section \ref{sec:twoformat}.
The context generated here differs from the Environment Response; it often contains structured, noisy, or verbose text that the agent must parse, filter, and integrate into its working memory.

\subsection{Data Construction}
The dataset is constructed through automated environment rollouts governed by a rule-based simulation. The generation process begins with the random selection of an initial item to trigger the feedback loop. For subsequent rounds, we generate a sequence of [Tool use, Tool Response, Guess, Environment Response] for each interaction step. To achieve scalable context lengths, we parametrically adjust the granularity of puzzle constraints or sequentially chain related interaction sessions. This methodology ensures that the expanded context remains causally linked and logically coherent, avoiding the disjointed nature of random document concatenation.

\subsection{Task Description}
\label{sec:task_description}
We decompose the evaluation into specific cognitive dimensions to pinpoint the mechanisms of agentic failure.

\subsubsection{Two Settings}

We differentiate the evaluation based on the semantic nature of the information processed. The \textbf{Knowledge-Intensive} setting constructs trajectories using real-world entities, instantiated here with the Pokémon dataset. Although the evaluation answers are theoretically derivable solely from the context history, the presence of familiar entities triggers the model's parametric memory. This design mimics realistic domain-specific tasks where models may hallucinate based on prior knowledge—such as predicting a type based on a name rather than tool output—or conversely, benefit from domain familiarity.

To rigorously evaluate pure in-context reasoning, we employ a \textbf{Knowledge-Free} setting via fully symbolic masking. Unlike previous benchmarks that merely perform entity substitution (e.g., changing "Newton" to "John") while retaining semantic structures, we map all entities and attributes to abstract tokens. Specifically, item names are mapped to IDs (e.g., \texttt{Item\_84}), and attributes are mapped to abstract codes (e.g., \texttt{Attr\_1} with value \texttt{A1V1}). This approach eliminates semantic cues, forcing the model to rely exclusively on the logical constraints defined within the interaction history. This provides an unbiased measurement of state tracking and memory retention, free from the interference of pre-trained knowledge.

\subsubsection{Two Formats}
\label{sec:twoformat}
To investigate how information distribution impacts performance, we apply two distinct interaction formats. While both are controlled to maintain comparable total context lengths, they present fundamentally different challenges regarding information density.

The \textbf{Concise-Response} format prioritizes turns extension. Here, the tool returns only the pre-calculated intersection of candidate items that satisfy the queried attributes.To accumulate a total context length comparable to the Verbose format, this setting involves a massive number of interaction turns (e.g., hundreds of rounds). This design evaluates whether agents can maintain consistent state tracking over hundreds of rounds without losing early constraints.

Conversely, the \textbf{Verbose-Response} format tests the capacity to handle information overload. In this setting, tools return full, unfiltered candidate lists for each queried attribute. A query for three attributes results in three extensive lists, creating a trajectory with fewer turns but high-density blocks of structured text. The agent is thus forced to perform logical intersections internally and extract key information from massive, noisy inputs within a single step.




\subsubsection{Task Taxonomy}
\label{sec:tasks_taxonomy_summary}

To pinpoint the specific cognitive mechanisms behind agentic failures, we classify the eight tasks into three dimensions (visualized in Figure \ref{fig:methodology_overview}; see Appendix \ref{sec:appendix_tasks} for detailed definitions).

\textbf{QA in Tool Response} evaluates the robustness of parsing machine-generated logs. Tasks such as \textit{Find Duplicates} measure the ability to extract precise details from structured noise, explicitly testing resilience against information overload.

\textbf{QA in Environment Response} targets the capacity to track evolving states and interpret historical feedback constraints, exemplified by the \textit{Weighted Summation} task. This dimension assesses how well an agent maintains a coherent belief state across long, iterative interactions, isolating the impact of memory fragmentation.

\textbf{Final Guess (Intersection)} serves as the ultimate test of global understanding. It requires the agent to perform logical set operations across the entire context trajectory to deduce the target, thereby integrating retrieval with deductive reasoning.

This dimensional breakdown allows us to distinguish whether performance degradation stems from local retrieval errors, calculation deficits, or a fundamental breakdown in logical consistency.

\begin{figure*}[t]
    \centering
    \includegraphics[width=\linewidth]{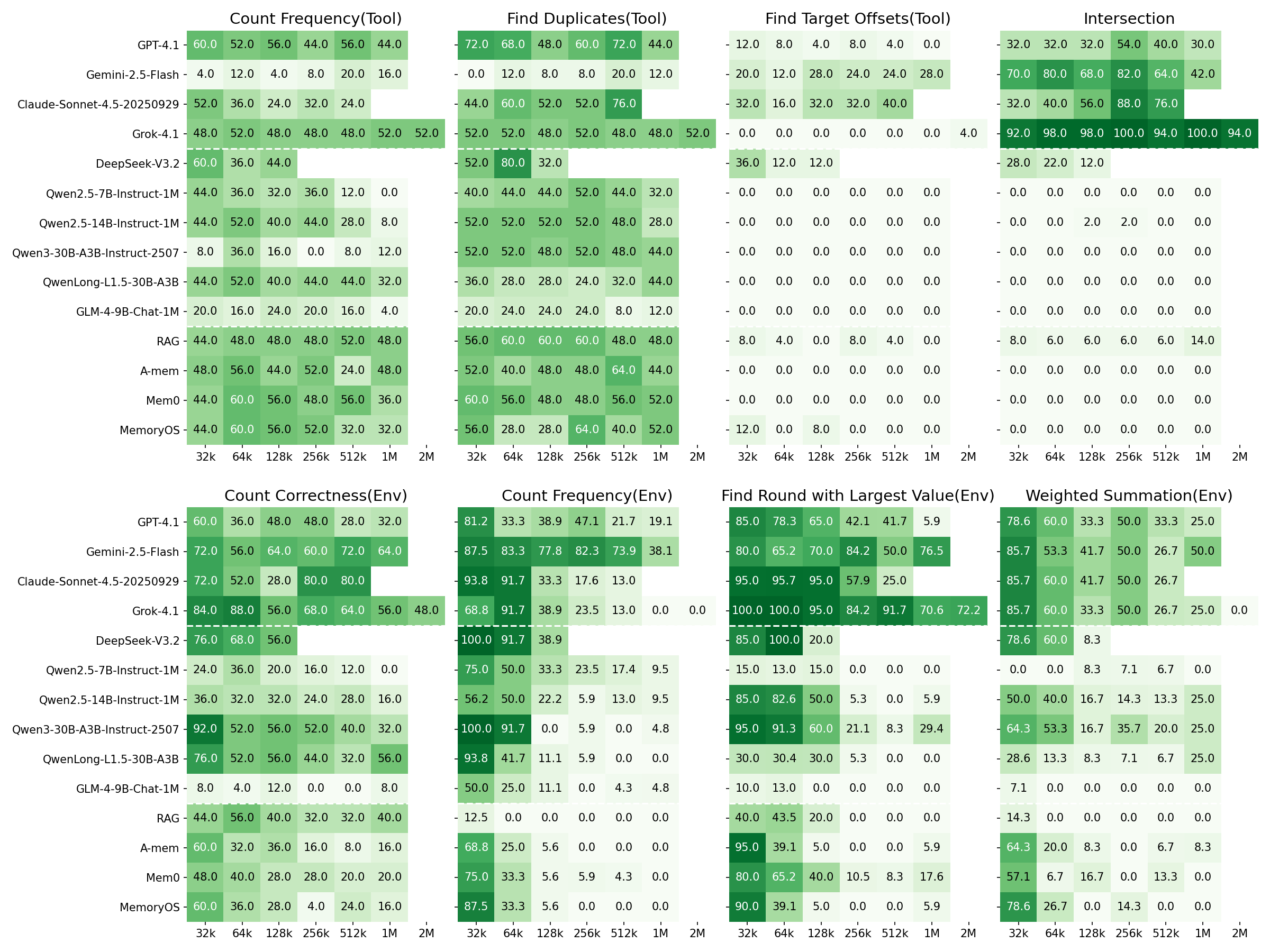}
    \caption{Main Results on Knowledge-Intensive \& Concise-Response Setting. The heatmap visualizes model performance across varying context lengths (32K to 2M). Green indicates higher accuracy. }
    \label{fig:exp_ki_concise}
\end{figure*}

\begin{figure*}[t]
    \centering
    \includegraphics[width=0.98\linewidth]{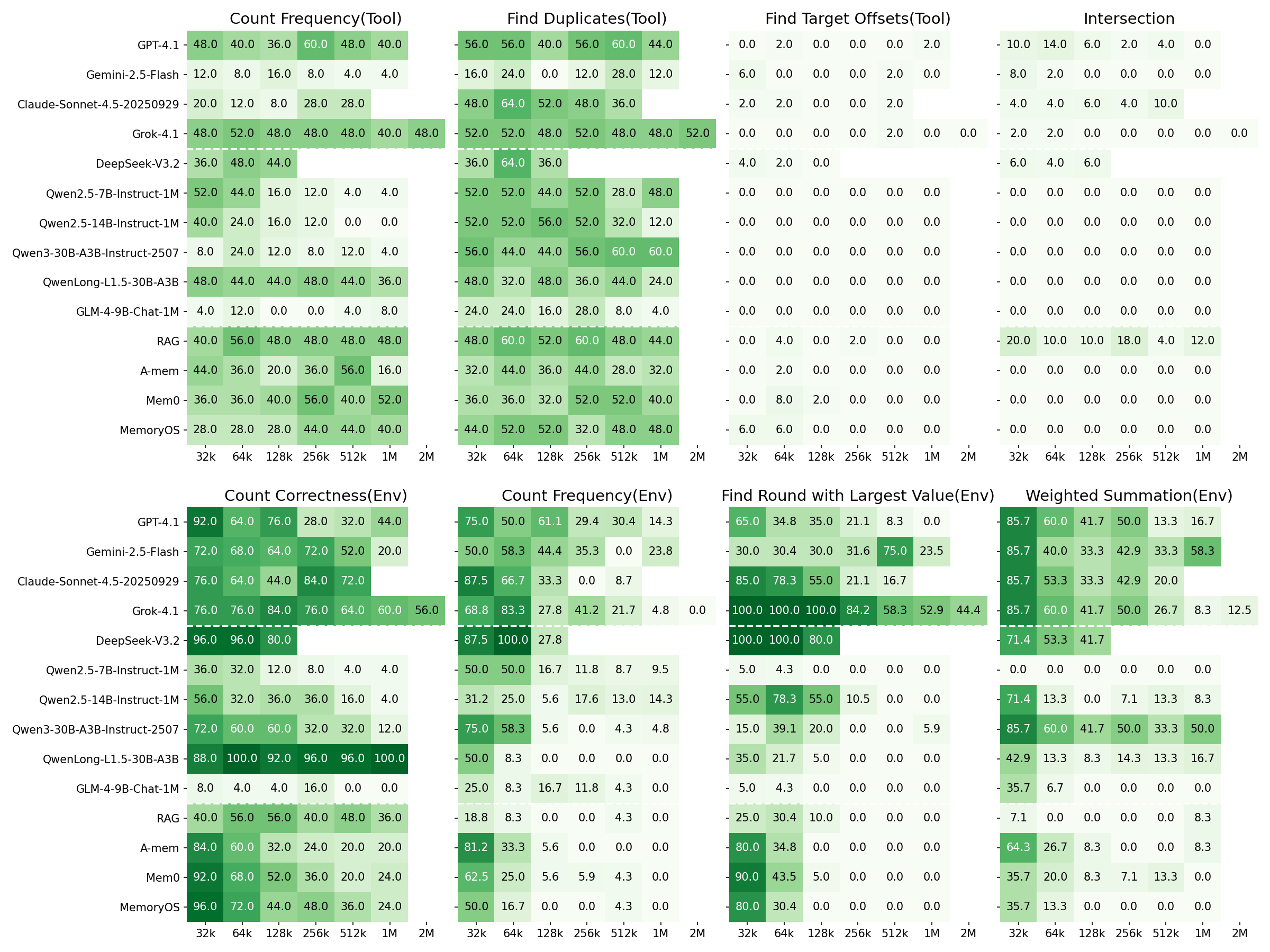}
    \caption{Main Results on Knowledge-Free \& Concise-Response Setting. }
    \label{fig:exp_kf_concise}
\end{figure*}

\begin{figure*}[h]
    \centering
    \includegraphics[width=\linewidth]{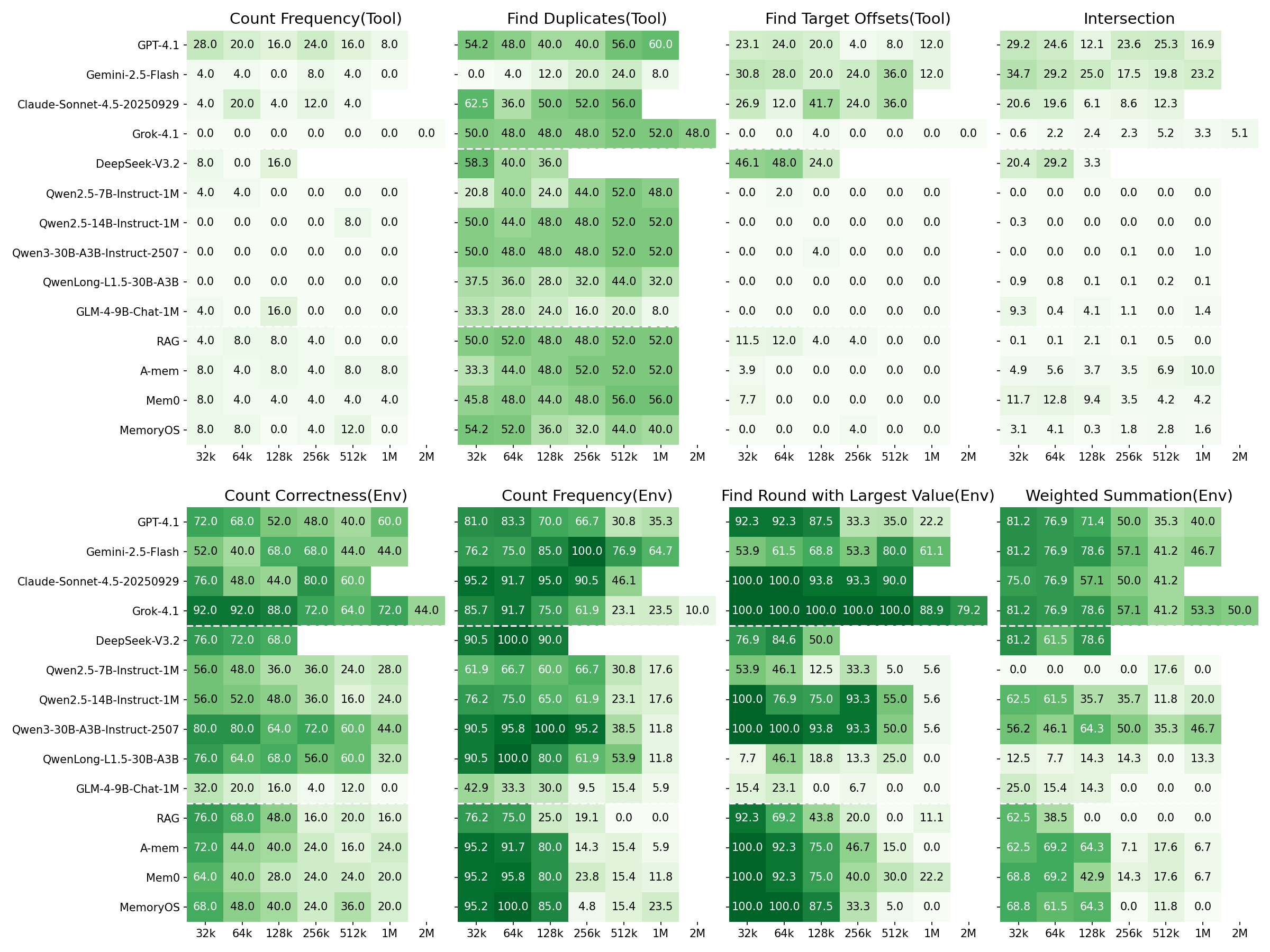}
    \caption{Main Results on Knowledge-Intensive \& Verbose-Response Setting.}
    \label{fig:exp_ki_verbose}
\end{figure*}

\begin{figure*}[h]
    \centering
    \includegraphics[width=\linewidth]{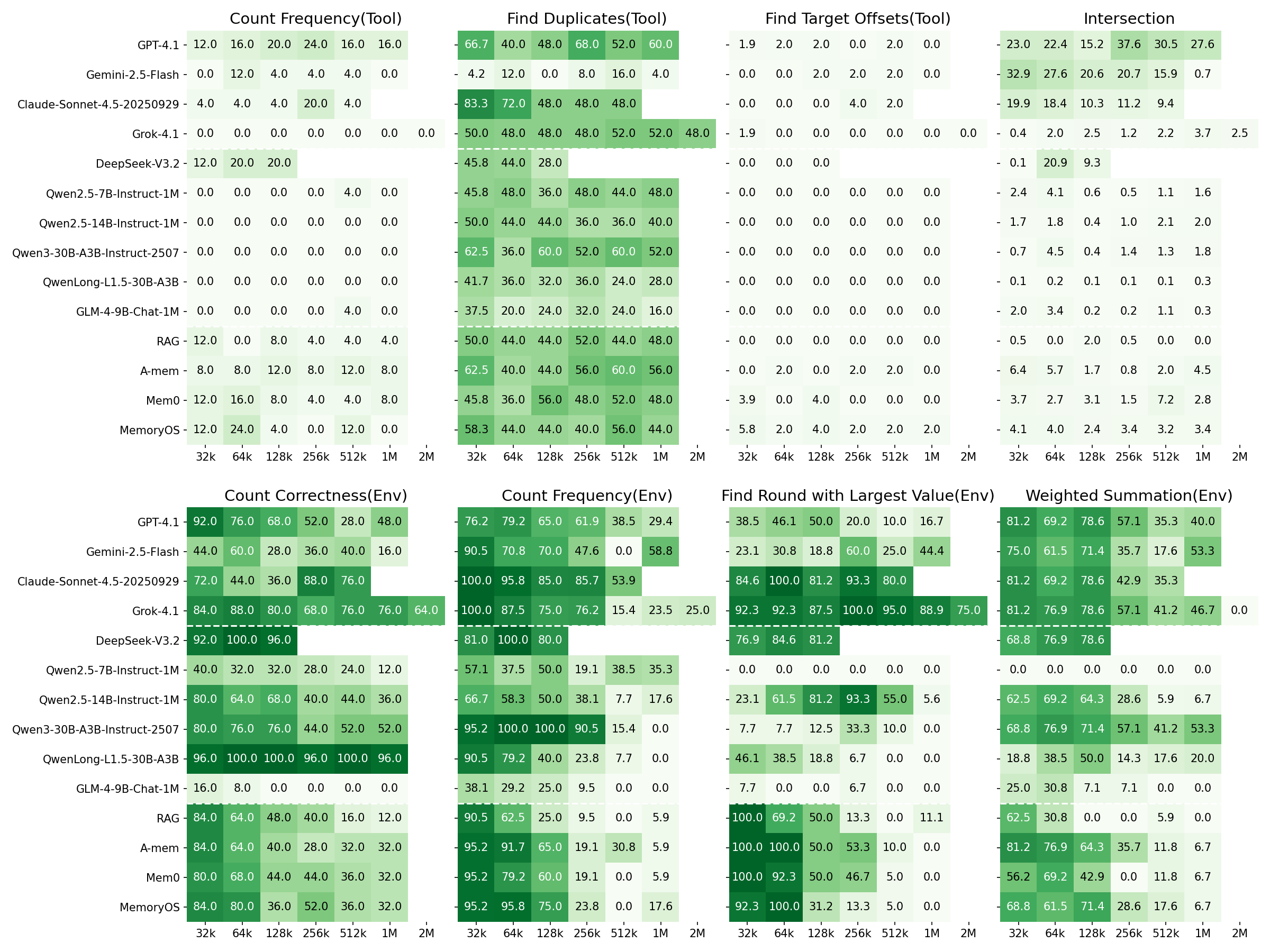}
    \caption{Main Results on Knowledge-Free \& Verbose-Response Setting.}
    \label{fig:exp_kf_verbose}
\end{figure*}

\section{Experiments}

\subsection{Implementations}
We evaluate a diverse array of state-of-the-art Large Language Models (LLMs), encompassing both proprietary and open-weight systems. On the proprietary front, our analysis includes GPT-4.1 \citep{openai_mrcr}, Gemini-2.5-Flash \citep{comanici2025gemini}, Claude-Sonnet-4.5 \citep{sonnet-4.5}, and Grok-4.1 \citep{grok-4.1}. These models are generally assessed at context lengths up to 1M tokens, with the exception of Grok-4.1, which is evaluated up to 2M tokens. For open-source models, we select representative long-context checkpoints, including DeepSeek-V3.2 \citep{liu2025deepseek}, the Qwen series (specifically Qwen2.5-7/14B \citep{yang2025qwen2} and Qwen3-30B-A3B \citep{yang2025qwen3}), QwenLong-L1.5-30B-A3B \citep{shen2025qwenlong}, and GLM-4-9B-Chat-1M \citep{glm2024chatglm}.

Beyond native long-context models, we also benchmark external memory architectures. We implement standard RAG \citep{lewis2020retrieval} alongside specialized agentic memory systems such as A-Mem \citep{a-mem}, Mem0 \citep{chhikara2025mem0buildingproductionreadyai}, and MemoryOS \citep{kang2025memoryosaiagent}. To control for reasoning variance and isolate the contribution of the memory mechanism, all such frameworks employ \textbf{Qwen3-30B-A3B-Instruct-2507} as a unified backbone.For reproducibility, we adhere to standard public configurations for these frameworks, with full implementation details provided in Appendix \ref{sec:baselines}.

\subsection{Main Evaluation Results}

Here, we present the performance heatmaps for all settings from Figures \ref{fig:exp_ki_concise} to \ref{fig:exp_kf_verbose}.

\begin{figure*}[t]
    \centering
    \begin{minipage}{0.48\textwidth}
        \centering
        \includegraphics[width=\linewidth]{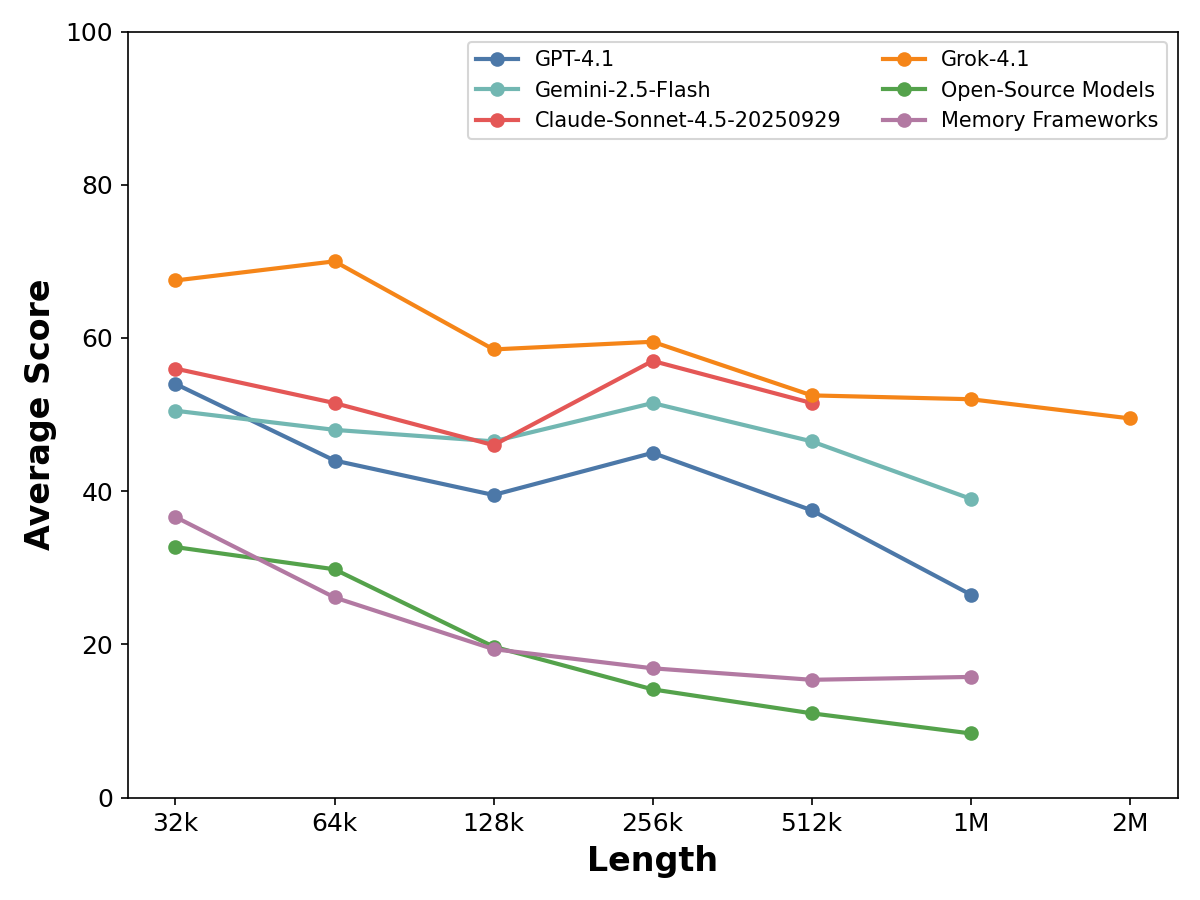} 
        \caption{Average score trends across context lengths. Performance comparison of Proprietary models, aggregated Open-Source models, and Memory Frameworks.}
        \label{fig:trend_models}
    \end{minipage}
    \hfill
    \begin{minipage}{0.48\textwidth}
        \centering
        \includegraphics[width=\linewidth]{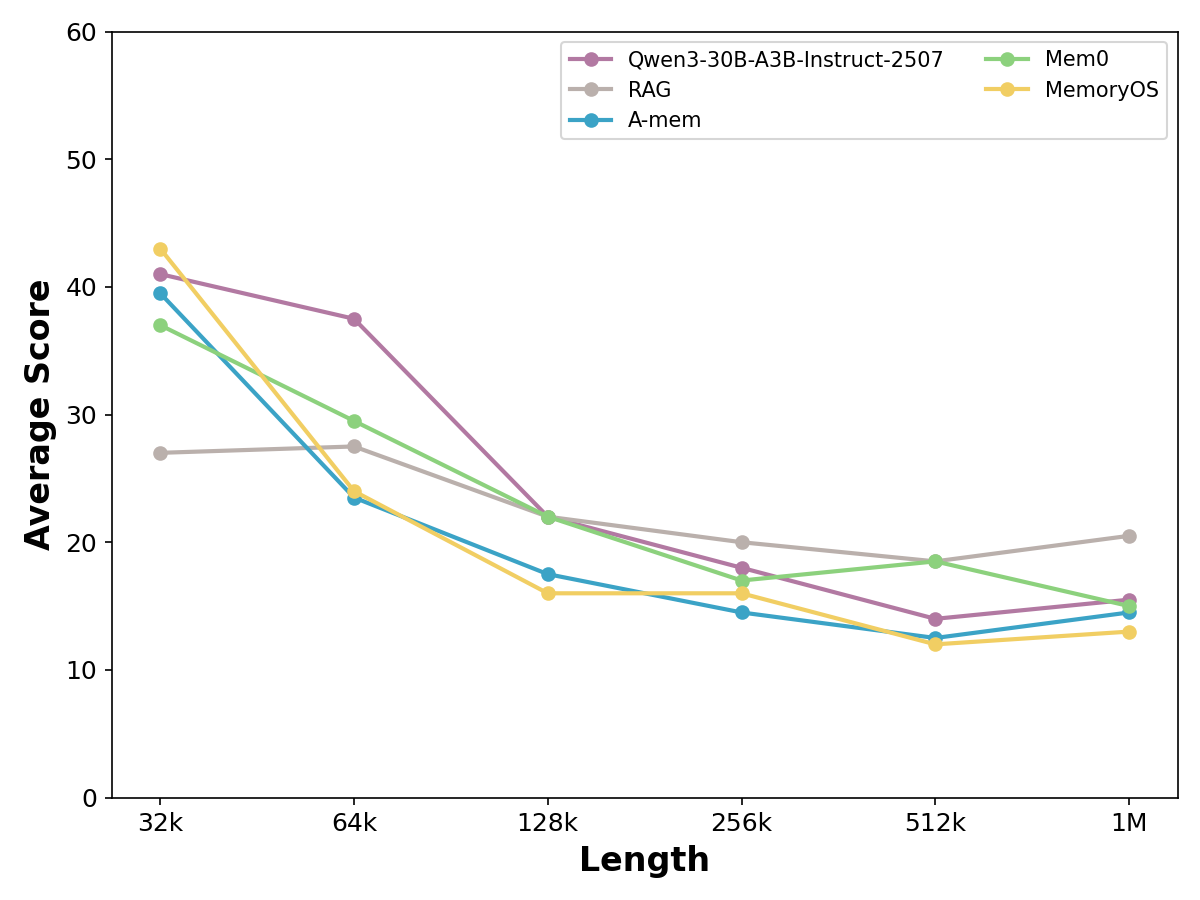}
        \caption{Comparison of Memory Frameworks vs. Base Model. Evaluation of RAG and specialized memory agents on the Qwen3-30B backbone.}
        \label{fig:trend_memory}
    \end{minipage}
\end{figure*}

\noindent\textbf{Model Performance across Context Lengths.} 
Figure \ref{fig:trend_models} reveals a distinct performance gap between proprietary frontiers and open-weight models (see Figures \ref{fig:exp_ki_concise} and \ref{fig:exp_kf_concise} for detailed heatmaps). Grok-4.1 demonstrates remarkable resilience, maintaining scores above 50.0 even as the context extends to 2M tokens. Conversely, while Gemini-2.5-Flash and GPT-4.1 display strong initial reasoning, their performance decays significantly beyond the 256k token threshold, dropping below 40.0 and 30.0 respectively at 1M tokens. Open-source models face an even steeper challenge: despite architectural improvements in DeepSeek-V3.2 and the Qwen3 series, aggregated performance starts at a lower baseline and degrades to negligible levels by 1M tokens.

\vspace{0.5em}
\noindent\textbf{High-Precision Task Constraints.} 
This degradation is particularly acute in tasks requiring strict positional awareness, such as \textit{Find Target Offsets}. Unlike semantic retrieval questions where approximate matches often suffice, these tasks demand that the agent precisely locate specific indices within dense, machine-generated logs. A single hallucinated offset or a missed entry in the history breaks the necessary logical chain. This zero-tolerance for errors explains why this task performs poorly.

\vspace{0.5em}
\noindent\textbf{Inefficacy of Memory Augmentation.} 
Turning to external memory mechanisms (Figure \ref{fig:trend_memory}), the results are counter-intuitive: specialized frameworks fail to improve upon the standalone Qwen3-30B-A3B-Instruct backbone. The base model consistently outperforms memory-augmented variants across most context lengths. Although MemoryOS achieves a marginal lead at short contexts (32k), it succumbs to rapid degradation as the interaction history grows, eventually falling behind the native context window. Standard RAG remains stable but plateaus at a significantly lower performance level.

These shortcomings likely stem from a misalignment between agentic data and retrieval logic. Agentic workflows generate highly structured tool outputs (e.g., JSON arrays) that confound generic extractors, which frequently introduce bias or fail to retrieve the complete set of constraints. Moreover, the process of compressing history into summaries or vector indices is fundamentally at odds with the requirements of lateral thinking puzzles. Since every historical constraint acts as a necessary premise for deduction, the "lossy" retrieval typical of RAG and memory agents severs the logical dependencies required to solve the puzzle, resulting in the observed deficit compared to full-context processing.

\section{Analysis}

This section analyzes the experimental results with an eye toward concrete failure modes. We organize the discussion around the three research questions in Section~\ref{sec:task_description}.

\subsection{Knowledge-Intensive vs.\ Knowledge-Free}

We first contrast the \textit{Knowledge-Intensive} and \textit{Knowledge-Free} settings (Figure~\ref{fig:exp_ki_concise} and Figure~\ref{fig:exp_kf_concise}). A clear example is the Intersection task. In the Knowledge-Intensive setting, models achieve non-trivial accuracy at shorter contexts (e.g., GPT-4.1 reaches roughly 30--40\%). In the Knowledge-Free setting, the same task drops to near-zero performance for nearly all models, including the strongest ones.

This gap suggests that the Knowledge-Intensive setting allows models to lean on parametric associations between familiar entities and attributes, which can partially substitute for the explicit set operations required by the context. When those semantic cues are removed, models must rely almost entirely on symbolic state tracking over the interaction history, where performance deteriorates sharply. In other words, the Knowledge-Free setting exposes a weakness in maintaining and updating discrete logical states under long-horizon interaction.

\subsection{Concise-Response vs.\ Verbose-Response}

We analyze how the response format reshapes difficulty by comparing the \textit{Concise-Response} setting (Figure~\ref{fig:exp_ki_concise}) with the \textit{Verbose-Response} setting (Figure~\ref{fig:exp_ki_verbose}). The key difference is not simply whether responses are ``more informative,'' but how information that is \emph{irrelevant to the queried target} is distributed across an episode.

For tasks that query Environment Response (e.g., \textit{Count Frequency (Env)}), the Verbose format often yields higher accuracy. This improvement does not come from richer tool outputs---the tool content is largely incidental for these queries. Instead, verbose responses reduce the number of interaction turns. Under the concise format, episodes become much longer, and the model must maintain the evolving state across many steps even though most tool-return tokens are not needed for answering. This makes long-horizon state tracking the dominant bottleneck, and performance degrades accordingly.

The trend reverses for tasks that query Tool Response(e.g., \textit{Find Duplicates}). Here the tool-return content is the evidence. The Verbose format concentrates a large amount of machine-generated text into fewer turns, which increases the density of distractors and the cost of locating the relevant fields within a single response. In contrast, the Concise format spreads tool outputs across more turns, but each turn is smaller and easier to scan, which can make evidence localization and comparison more manageable for these tasks.

Overall, the two formats trade off where the difficulty lies: Concise increases episode length and long-range state maintenance, while Verbose increases within-turn density when the answer must be extracted from tool logs.

\subsection{Environment Response vs.\ Tool Response and Adequate Context Length}
Across all four settings, we observe a consistent gap between tasks that query Tool Response and those that query Environment Response. In Figure~\ref{fig:exp_ki_concise}, tool-response tasks (top row) are uniformly harder than environment-response tasks (bottom row), even when the total context length is matched.

To account for this gap, we introduce \textbf{Adequate Context Length (ACL)}. ACL is a property of the \emph{input trajectory}: it measures how many tokens a model must traverse to locate and assemble the evidence needed for a single query. Importantly, ACL is computed from the episode text alone and does not depend on model outputs or prediction correctness.

For environment-response tasks such as \textit{Count Frequency (Env)}, models generally favor the Verbose format. This advantage stems from the experimental design rather than the content itself: Verbose trajectories inherently comprise significantly fewer interaction steps than their Concise counterparts. Crucially, since the answers to these tasks are derived solely from environmental feedback, the voluminous tool outputs in the Verbose format serve merely as distractors which do not aid reasoning. The results suggest that models find it easier to bypass this dense, irrelevant noise within a single turn than to maintain a coherent state across the hundreds of fragmented turns characteristic of the Concise format.

The pattern reverses for tasks that query tool logs, such as QA in Tool Response (e.g., \textit{Find Duplicates}). Here, Verbose responses often hurt. Consolidating tool outputs into a single turn yields long, dense, machine-generated text where relevant fields are mixed with large amounts of irrelevant structure. This creates information overload: the model must extract and align specific values from the verbose block.

Table~\ref{tab:acl_comparison} quantifies this difference. Even with identical total context lengths, tool-response tasks impose a much larger ACL, which increases the difficulty of evidence localization and aggregation and is associated with substantially lower accuracy.

\begin{table}[h]
    \centering
    \small
    \begin{tabular}{l|l|c|c}
    \toprule
        \textbf{Format} & \textbf{Query target} & \textbf{ACL (tokens)} & \textbf{Acc.(128k)} \\
        \midrule
        \multirow{2}{*}{Concise} & Env response & 2044.1 & 47.3\% \\
         & Tool response & 3040.8 & 36.0\% \\
        \midrule
        \multirow{2}{*}{Verbose} & Env response & 535.8 & 68.2\% \\
         & Tool response & 11439.6 & 25.3\% \\
        \bottomrule
    \end{tabular}    
    \caption{Performance and adequate context length. We report GPT-4.1 accuracy at a 128k context window. ACL is measured in tokens and computed from the input trajectory only, independent of model outputs. Within each format, tool-response queries have substantially larger ACL and lower accuracy than environment-response queries, consistent with the intuition that evidence localization becomes harder as the required span grows.}
    \label{tab:acl_comparison}
\end{table}

\section{Related works}
\label{appendix:relatedworks}

\textbf{Benchmarks for Long Contexts.} The evaluation of long-context LLMs has evolved from adapting traditional NLP tasks to designing complex, synthetic reasoning scenarios. However, the majority of these efforts focus on the static context with no close relation with one target, showing less reality in agent scenerios..

\textit{General Understanding and Retrieval.} 
Initial benchmarks, such as \citep{shaham2022scrolls,shaham2023zeroscrolls} and \citep{bai2023longbench}, established the foundation by aggregating datasets like NarrativeQA~\citep{kovcisky2018narrativeqa} and GovReport~\citep{huang2021efficient}. These were further standardized by suites like L-Eval~\citep{an2024eval} and LongBenchV2~\citep{bai2024longbench2} to assess general capabilities across varying lengths. 
A major focus has been strictly testing retrieval limits, popularized by the "Needle-In-A-Haystack" paradigm~\citep{niah}. This approach has expanded into comprehensive suites like RULER~\citep{hsiehruler} and NeedleBench~\citep{li2024needlebench}, covering multi-needle retrieval~\citep{reid2024gemini,openai_mrcr}, citation evaluation~\citep{yen2025helmet,li2024scbench}, and instruction following stability~\citep{wu2025lifbench}. Recent studies have also utilized these setups to investigate phenomena like "Lost-In-the-Middle"~\citep{liu2024lost,goldman2024really} and the impact of demonstration quality in long in-context learning~\citep{li2024long,agarwal2024many,bertsch2024context,wang2024precision,kai2025mirbench}.

\textit{Reasoning, Logic, and Domain Specifics.}
To evaluate deeper cognitive functions beyond passive retrieval, recent works have introduced tasks requiring logic, mathematics, and code understanding. $\infty$-Bench~\citep{zhang2024bench} and BAMBOO~\citep{dong2024bamboo} target complex reasoning across diverse domains, while others focus on repository-level code~\citep{liu2024repoqa} and mathematical noise filtering~\citep{wang2024mathhay}. 
A critical direction involves testing long-range dependencies and robustness. Benchmarks like LooGLE~\citep{li2024loogle,he2025loogle} and BABILong~\citep{NEURIPS2024_babilong} challenge models to trace multi-hop evidence or comprehend native long texts such as novels~\citep{wang2024novelqa,xu2024detectiveqa,ye2025longproc}. Concurrently, works like LV-Eval~\citep{yuan2024lv} and NoCha~\citep{Karpinska2024OneTA} probe model robustness against confounding facts. 
Other efforts emphasize data synthesis and controllability: Michelangelo~\citep{Vodrahalli2024MichelangeloLC} and HoloBench~\citep{maekawa2024holistic} test structure discernment, while Ada-LEval~\citep{wang2024ada} and LongBioBench~\citep{yang2025controllable} utilize generated content to decouple reasoning from parametric memory. Loong~\citep{wang2024leave} further pushes density by ensuring every document is crucial.

\textit{Agentic and Memory Evaluations.}
As LLMs evolve into agents, evaluating their ability to maintain history is critical. LocoMo~\citep{maharana2024evaluating} and LongMemEval~\citep{wu2024longmemeval} assess memory retention across long conversational histories, while MemoryAgentBench~\citep{hu2025evaluating} reconstructs long benches and focuses on memory recall accuracy. 

\textbf{Long LLMs.} Recent advancements in large language models have substantially extended their context window, with state-of-the-art models claiming to support up to 128K or even 2M tokens~\citep{openai_mrcr,comanici2025gemini,sonnet-4.5,guo2025deepseek,liu2024deepseek,grok-4.1}. Meanwhile, various efforts have been made to extend models' context length and enhance their long-dependency capabilities. These include more efficient attention mechanisms~\citep{dao2023flashattention,xiao2024duoattention,yuan2025native}, scalable training strategies such as test-time training and parameter-efficient fine-tuning~\citep{sun2020test,chen2023longlora}, and length-extrapolatable positional encodings~\citep{su2024roformer,pengyarn,ding2024longrope}. Together, these innovations reduce computational overhead while preserving the model’s ability to retain distant information, thus enabling more effective reasoning over extended contexts.

\textbf{RAG and Memory Systems.} The concept of augmenting language models with external knowledge retrieval has gained significant traction. Early work on Dense Passage Retrieval (DPR)~\citep{karpukhin2020dense} demonstrated the effectiveness of dense vector representations, leading to the proposal of Retrieval-Augmented Generation (RAG)~\citep{lewis2020retrieval} and subsequent developments~\citep{alce, udr}. To enhance retrieval quality, LightRAG~\citep{LightRAG} employs a dual-level system, while structure-based methods like GraphRAG~\citep{GraphRAG} and others~\citep{PathRAG,HippoRAG2,HyperGraphRAG} utilize fine-grained entities or links. \textbf{Agentic Memory Agents} introduce an iterative, decision-driven framework. Unlike single-pass retrieval, these agents dynamically process queries through multiple reasoning cycles. Examples include MemGPT~\citep{MemGPT} and other recent systems~\citep{a-mem,chhikara2025mem0buildingproductionreadyai,kang2025memoryosaiagent,mem1,MemAgent,mirix}.

\section{Conclusion}

We introduced \textbf{AgentLongBench}, a long-context benchmark built around agent--environment interaction rather than static document retrieval. Using automated rollouts, AgentLongBench generates verifiable trajectories at scale and supports evaluation over extended horizons.

Experiments on frontier LLMs and memory-augmented agents reveal a sharp drop in accuracy as episodes grow longer and tool use becomes more involved. The failures are not explained by context length alone. Models often rely on parametric shortcuts when semantic cues are available, but struggle when they must track and update discrete states over many steps. They also degrade when the answer is embedded in dense tool logs, where extracting and aligning the relevant fields becomes difficult even at the same total context length. Overall, current RAG and memory mechanisms do not reliably support long-horizon state tracking and high-ACL evidence localization, suggesting that robust tool-grounded reasoning remains an open bottleneck for long-context agents.

\clearpage
\bibliographystyle{plainnat}
\bibliography{custom}

@misc{kang2025memoryosaiagent,
      title={Memory OS of AI Agent}, 
      author={Jiazheng Kang and Mingming Ji and Zhe Zhao and Ting Bai},
      year={2025},
      eprint={2506.06326},
      archivePrefix={arXiv},
      primaryClass={cs.AI},
      url={https://arxiv.org/abs/2506.06326}, 
}

@misc{chhikara2025mem0buildingproductionreadyai,
      title={Mem0: Building Production-Ready AI Agents with Scalable Long-Term Memory}, 
      author={Prateek Chhikara and Dev Khant and Saket Aryan and Taranjeet Singh and Deshraj Yadav},
      year={2025},
      eprint={2504.19413},
      archivePrefix={arXiv},
      primaryClass={cs.CL},
      url={https://arxiv.org/abs/2504.19413}, 
}

@misc{pokemon,
      title={Pokemon Wikipedia}, 
      author={},
        url={https://en.wikipedia.org/wiki/Pok%C3%A9mon}
}

@misc{oxfordlateralthinking,
      title={Oxford English Dictionary: Lateral thinking}, 
      author={},
      year={2016},
        publisher={Oxford University Press}
}

@article{lewis2020retrieval,
  title={Retrieval-augmented generation for knowledge-intensive nlp tasks},
  author={Lewis, Patrick and Perez, Ethan and Piktus, Aleksandra and Petroni, Fabio and Karpukhin, Vladimir and Goyal, Naman and K{\"u}ttler, Heinrich and Lewis, Mike and Yih, Wen-tau and Rockt{\"a}schel, Tim and others},
  journal={Advances in neural information processing systems},
  volume={33},
  pages={9459--9474},
  year={2020}
}

@misc{GraphRAG,
      title={From Local to Global: A Graph RAG Approach to Query-Focused Summarization}, 
      author={Darren Edge and Ha Trinh and Newman Cheng and Joshua Bradley and Alex Chao and Apurva Mody and Steven Truitt and Dasha Metropolitansky and Robert Osazuwa Ness and Jonathan Larson},
      year={2025},
      eprint={2404.16130},
      archivePrefix={arXiv},
      primaryClass={cs.CL},
      url={https://arxiv.org/abs/2404.16130}, 
}

@misc{LightRAG,
      title={LightRAG: Simple and Fast Retrieval-Augmented Generation}, 
      author={Zirui Guo and Lianghao Xia and Yanhua Yu and Tu Ao and Chao Huang},
      year={2025},
      eprint={2410.05779},
      archivePrefix={arXiv},
      primaryClass={cs.IR},
      url={https://arxiv.org/abs/2410.05779}, 
}

@misc{HyperGraphRAG,
      title={HyperGraphRAG: Retrieval-Augmented Generation via Hypergraph-Structured Knowledge Representation}, 
      author={Haoran Luo and Haihong E and Guanting Chen and Yandan Zheng and Xiaobao Wu and Yikai Guo and Qika Lin and Yu Feng and Zemin Kuang and Meina Song and Yifan Zhu and Luu Anh Tuan},
      year={2025},
      eprint={2503.21322},
      archivePrefix={arXiv},
      primaryClass={cs.AI},
      url={https://arxiv.org/abs/2503.21322}, 
}

@misc{PathRAG,
      title={PathRAG: Pruning Graph-based Retrieval Augmented Generation with Relational Paths}, 
      author={Boyu Chen and Zirui Guo and Zidan Yang and Yuluo Chen and Junze Chen and Zhenghao Liu and Chuan Shi and Cheng Yang},
      year={2025},
      eprint={2502.14902},
      archivePrefix={arXiv},
      primaryClass={cs.CL},
      url={https://arxiv.org/abs/2502.14902}, 
}

@misc{HippoRAG2,
      title={From RAG to Memory: Non-Parametric Continual Learning for Large Language Models}, 
      author={Bernal Jiménez Gutiérrez and Yiheng Shu and Weijian Qi and Sizhe Zhou and Yu Su},
      year={2025},
      eprint={2502.14802},
      archivePrefix={arXiv},
      primaryClass={cs.CL},
      url={https://arxiv.org/abs/2502.14802}, 
}

@inproceedings{karpukhin2020dense,
  title={Dense Passage Retrieval for Open-Domain Question Answering.},
  author={Karpukhin, Vladimir and Oguz, Barlas and Min, Sewon and Lewis, Patrick SH and Wu, Ledell and Edunov, Sergey and Chen, Danqi and Yih, Wen-tau},
  booktitle={EMNLP (1)},
  pages={6769--6781},
  year={2020}
}

@article{guo2025deepseek,
  title={Deepseek-r1: Incentivizing reasoning capability in llms via reinforcement learning},
  author={Guo, Daya and Yang, Dejian and Zhang, Haowei and Song, Junxiao and Zhang, Ruoyu and Xu, Runxin and Zhu, Qihao and Ma, Shirong and Wang, Peiyi and Bi, Xiao and others},
  journal={arXiv preprint arXiv:2501.12948},
  year={2025}
}

@inproceedings{alce,
   title={Enabling Large Language Models to Generate Text with Citations},
   author={Gao, Tianyu and Yen, Howard and Yu, Jiatong and Chen, Danqi},
   year={2023},
   booktitle={Empirical Methods in Natural Language Processing (EMNLP)},
}

@inproceedings{udr,
    title = "Unified Demonstration Retriever for In-Context Learning",
    author = "Li, Xiaonan  and
      Lv, Kai  and
      Yan, Hang  and
      Lin, Tianyang  and
      Zhu, Wei  and
      Ni, Yuan  and
      Xie, Guotong  and
      Wang, Xiaoling  and
      Qiu, Xipeng",
    editor = "Rogers, Anna  and
      Boyd-Graber, Jordan  and
      Okazaki, Naoaki",
    booktitle = "Proceedings of the 61st Annual Meeting of the Association for Computational Linguistics (Volume 1: Long Papers)",
    month = jul,
    year = "2023",
    address = "Toronto, Canada",
    publisher = "Association for Computational Linguistics",
    url = "https://aclanthology.org/2023.acl-long.256/",
    doi = "10.18653/v1/2023.acl-long.256",
    pages = "4644--4668"
}

@article{he2025loogle,
  title={LooGLE v2: Are LLMs Ready for Real World Long Dependency Challenges?},
  author={He, Ziyuan and Wang, Yuxuan and Li, Jiaqi and Liang, Kexin and Zhang, Muhan},
  journal={arXiv preprint arXiv:2510.22548},
  year={2025}
}

@article{hu2025evaluating,
  title={Evaluating memory in llm agents via incremental multi-turn interactions},
  author={Hu, Yuanzhe and Wang, Yu and McAuley, Julian},
  journal={arXiv preprint arXiv:2507.05257},
  year={2025}
}

@article{MemGPT,
  title={MemGPT: Towards LLMs as Operating Systems.},
  author={Packer, Charles and Fang, Vivian and Patil, Shishir\_G and Lin, Kevin and Wooders, Sarah and Gonzalez, Joseph\_E},
  year={2023},
  publisher={arXiv}
}

@article{a-mem,
  title={A-mem: Agentic memory for llm agents},
  author={Xu, Wujiang and Mei, Kai and Gao, Hang and Tan, Juntao and Liang, Zujie and Zhang, Yongfeng},
  journal={arXiv preprint arXiv:2502.12110},
  year={2025}
}

@article{MEM1,
  title={MEM1: Learning to Synergize Memory and Reasoning for Efficient Long-Horizon Agents},
  author={Zhou, Zijian and Qu, Ao and Wu, Zhaoxuan and Kim, Sunghwan and Prakash, Alok and Rus, Daniela and Zhao, Jinhua and Low, Bryan Kian Hsiang and Liang, Paul Pu},
  journal={arXiv preprint arXiv:2506.15841},
  year={2025}
}

@article{MemAgent,
  title={MemAgent: Reshaping Long-Context LLM with Multi-Conv RL-based Memory Agent},
  author={Yu, Hongli and Chen, Tinghong and Feng, Jiangtao and Chen, Jiangjie and Dai, Weinan and Yu, Qiying and Zhang, Ya-Qin and Ma, Wei-Ying and Liu, Jingjing and Wang, Mingxuan and others},
  journal={arXiv preprint arXiv:2507.02259},
  year={2025}
}

@article{mirix,
  title={MIRIX: Multi-Agent Memory System for LLM-Based Agents},
  author={Wang, Yu and Chen, Xi},
  journal={arXiv preprint arXiv:2507.07957},
  year={2025}
}

@article{wang2024ada,
  title={Ada-leval: Evaluating long-context llms with length-adaptable benchmarks},
  author={Wang, Chonghua and Duan, Haodong and Zhang, Songyang and Lin, Dahua and Chen, Kai},
  journal={arXiv preprint arXiv:2404.06480},
  year={2024}
}

@inproceedings{wu2025lifbench,
  title={Lifbench: Evaluating the instruction following performance and stability of large language models in long-context scenarios},
  author={Wu, Xiaodong and Wang, Minhao and Liu, Yichen and Shi, Xiaoming and Yan, He and Xiangju, Lu and Zhu, Junmin and Zhang, Wei},
  booktitle={Proceedings of the 63rd Annual Meeting of the Association for Computational Linguistics (Volume 1: Long Papers)},
  pages={16445--16468},
  year={2025}
}

@article{yuan2024lv,
  title={Lv-eval: A balanced long-context benchmark with 5 length levels up to 256k},
  author={Yuan, Tao and Ning, Xuefei and Zhou, Dong and Yang, Zhijie and Li, Shiyao and Zhuang, Minghui and Tan, Zheyue and Yao, Zhuyu and Lin, Dahua and Li, Boxun and others},
  journal={arXiv preprint arXiv:2402.05136},
  year={2024}
}

@article{maekawa2024holistic,
  title={Holistic Reasoning with Long-Context LMs: A Benchmark for Database Operations on Massive Textual Data},
  author={Maekawa, Seiji and Iso, Hayate and Bhutani, Nikita},
  journal={arXiv preprint arXiv:2410.11996},
  year={2024}
}

@article{wang2024mathhay,
  title={Mathhay: An automated benchmark for long-context mathematical reasoning in llms},
  author={Wang, Lei and Dong, Shan and Xu, Yuhui and Dong, Hanze and Wang, Yalu and Saha, Amrita and Lim, Ee-Peng and Xiong, Caiming and Sahoo, Doyen},
  journal={arXiv preprint arXiv:2410.04698},
  year={2024}
}

@inproceedings{wang2024leave,
  title={Leave no document behind: Benchmarking long-context llms with extended multi-doc qa},
  author={Wang, Minzheng and Chen, Longze and Cheng, Fu and Liao, Shengyi and Zhang, Xinghua and Wu, Bingli and Yu, Haiyang and Xu, Nan and Zhang, Lei and Luo, Run and others},
  booktitle={Proceedings of the 2024 Conference on Empirical Methods in Natural Language Processing},
  pages={5627--5646},
  year={2024}
}

@inproceedings{zhang2024bench,
  title={$\infty$ Bench: Extending long context evaluation beyond 100k tokens},
  author={Zhang, Xinrong and Chen, Yingfa and Hu, Shengding and Xu, Zihang and Chen, Junhao and Hao, Moo and Han, Xu and Thai, Zhen and Wang, Shuo and Liu, Zhiyuan and others},
  booktitle={Proceedings of the 62nd Annual Meeting of the Association for Computational Linguistics (Volume 1: Long Papers)},
  pages={15262--15277},
  year={2024}
}

@inproceedings{li2024loogle,
  title={Loogle: Can long-context language models understand long contexts?},
  author={Li, Jiaqi and Wang, Mengmeng and Zheng, Zilong and Zhang, Muhan},
  booktitle={Proceedings of the 62nd Annual Meeting of the Association for Computational Linguistics (Volume 1: Long Papers)},
  pages={16304--16333},
  year={2024}
}

@article{maharana2024evaluating,
  title={Evaluating very long-term conversational memory of llm agents},
  author={Maharana, Adyasha and Lee, Dong-Ho and Tulyakov, Sergey and Bansal, Mohit and Barbieri, Francesco and Fang, Yuwei},
  journal={arXiv preprint arXiv:2402.17753},
  year={2024}
}

@inproceedings{wu2024longmemeval,
  title={A Benchmark for Long-Form Medical Question Answering},
  author={Hosseini, Pedram and Sin, Jessica M and Ren, Bing and Thomas, Bryceton G and Nouri, Elnaz and Farahanchi, Ali and Hassanpour, Saeed},
  booktitle={Advancements In Medical Foundation Models: Explainability, Robustness, Security, and Beyond}, 
  year={2024}
}

@article{comanici2025gemini,
  title={Gemini 2.5: Pushing the frontier with advanced reasoning, multimodality, long context, and next generation agentic capabilities},
  author={Comanici, Gheorghe and Bieber, Eric and Schaekermann, Mike and Pasupat, Ice and Sachdeva, Noveen and Dhillon, Inderjit and Blistein, Marcel and Ram, Ori and Zhang, Dan and Rosen, Evan and others},
  journal={arXiv preprint arXiv:2507.06261},
  year={2025}
}

@article{liu2025deepseek,
  title={Deepseek-v3. 2: Pushing the frontier of open large language models},
  author={Liu, Aixin and Mei, Aoxue and Lin, Bangcai and Xue, Bing and Wang, Bingxuan and Xu, Bingzheng and Wu, Bochao and Zhang, Bowei and Lin, Chaofan and Dong, Chen and others},
  journal={arXiv preprint arXiv:2512.02556},
  year={2025}
}

@misc{sonnet-4.5,
  title={Introducing Claude Sonnet 4.5},
  author={Anthropic },
  year={2025},
  url = {https://www.anthropic.com/news/claude-sonnet-4-5l}
}

@misc{grok-4.1,
  title={Grok 4.1 Model Card},
  author={xAI},
  year={2025},
  url = {https://data.x.ai/2025-11-17-grok-4-1-model-card.pdf}
}

@article{yang2025qwen2,
  title={Qwen2. 5-1m technical report},
  author={Yang, An and Yu, Bowen and Li, Chengyuan and Liu, Dayiheng and Huang, Fei and Huang, Haoyan and Jiang, Jiandong and Tu, Jianhong and Zhang, Jianwei and Zhou, Jingren and others},
  journal={arXiv preprint arXiv:2501.15383},
  year={2025}
}

@article{yang2025qwen3,
  title={Qwen3 technical report},
  author={Yang, An and Li, Anfeng and Yang, Baosong and Zhang, Beichen and Hui, Binyuan and Zheng, Bo and Yu, Bowen and Gao, Chang and Huang, Chengen and Lv, Chenxu and others},
  journal={arXiv preprint arXiv:2505.09388},
  year={2025}
}

@article{shen2025qwenlong,
  title={QwenLong-L1. 5: Post-Training Recipe for Long-Context Reasoning and Memory Management},
  author={Shen, Weizhou and Yang, Ziyi and Li, Chenliang and Lu, Zhiyuan and Peng, Miao and Sun, Huashan and Shi, Yingcheng and Liao, Shengyi and Lai, Shaopeng and Zhang, Bo and others},
  journal={arXiv preprint arXiv:2512.12967},
  year={2025}
}

@article{glm2024chatglm,
  title={Chatglm: A family of large language models from glm-130b to glm-4 all tools},
  author={GLM, Team and Zeng, Aohan and Xu, Bin and Wang, Bowen and Zhang, Chenhui and Yin, Da and Zhang, Dan and Rojas, Diego and Feng, Guanyu and Zhao, Hanlin and others},
  journal={arXiv preprint arXiv:2406.12793},
  year={2024}
}

@article{reid2024gemini,
  title={Gemini 1.5: Unlocking multimodal understanding across millions of tokens of context},
  author={Reid, Machel and Savinov, Nikolay and Teplyashin, Denis and Lepikhin, Dmitry and Lillicrap, Timothy and Alayrac, Jean-baptiste and Soricut, Radu and Lazaridou, Angeliki and Firat, Orhan and Schrittwieser, Julian and others},
  journal={arXiv preprint arXiv:2403.05530},
  year={2024}
}

@article{su2024roformer,
  title={Roformer: Enhanced transformer with rotary position embedding},
  author={Su, Jianlin and Ahmed, Murtadha and Lu, Yu and Pan, Shengfeng and Bo, Wen and Liu, Yunfeng},
  journal={Neurocomputing},
  volume={568},
  pages={127063},
  year={2024},
  publisher={Elsevier}
}

@inproceedings{pengyarn,
  title={YaRN: Efficient Context Window Extension of Large Language Models},
  author={Peng, Bowen and Quesnelle, Jeffrey and Fan, Honglu and Shippole, Enrico},
  booktitle={The Twelfth International Conference on Learning Representations},
  year={2024}
}

@article{ding2024longrope,
  title={Longrope: Extending llm context window beyond 2 million tokens},
  author={Ding, Yiran and Zhang, Li Lyna and Zhang, Chengruidong and Xu, Yuanyuan and Shang, Ning and Xu, Jiahang and Yang, Fan and Yang, Mao},
  journal={arXiv preprint arXiv:2402.13753},
  year={2024}
}

@article{xiao2024duoattention,
 author = {Xiao, Guangxuan and Tang, Jiaming and Zuo, Jingwei and Guo, Junxian and Yang, Shang and Tang, Haotian and Fu, Yao and Han, Song},
 journal = {arXiv preprint arXiv:2410.10819},
 year = {2024},
 title = {Duoattention: Efficient long-context llm inference with retrieval and streaming heads},
 venue = {arXiv preprint arXiv …}
}

@article{liu2024deepseek,
 author = {Liu, Aixin and Feng, Bei and Wang, Bin and Wang, Bingxuan and Liu, Bo and Zhao, Chenggang and Dengr, Chengqi and Ruan, Chong and Dai, Damai and Guo, Daya and others},
 journal = {arXiv preprint arXiv:2405.04434},
 year = {2024},
 title = {Deepseek-v2: A strong, economical, and efficient mixture-of-experts language model},
 venue = {arXiv preprint arXiv …}
}

@article{yuan2025native,
  title={Native Sparse Attention: Hardware-Aligned and Natively Trainable Sparse Attention},
  author={Jingyang Yuan and Huazuo Gao and Damai Dai and Junyu Luo and Liang Zhao and Zhengyan Zhang and Zhenda Xie and Y. X. Wei and Lean Wang and Zhiping Xiao and Yuqing Wang and Chong Ruan and Ming Zhang and Wenfeng Liang and Wangding Zeng},
  journal={arXiv preprint arXiv:2502.11089},
  year={2025}
}

@inproceedings{sun2020test,
  title={Test-time training with self-supervision for generalization under distribution shifts},
  author={Sun, Yu and Wang, Xiaolong and Liu, Zhuang and Miller, John and Efros, Alexei and Hardt, Moritz},
  booktitle={International conference on machine learning},
  pages={9229--9248},
  year={2020},
  organization={PMLR}
}

@article{kovcisky2018narrativeqa,
  title={The narrativeqa reading comprehension challenge},
  author={Ko{\v{c}}isk{\`y}, Tom{\'a}{\v{s}} and Schwarz, Jonathan and Blunsom, Phil and Dyer, Chris and Hermann, Karl Moritz and Melis, G{\'a}bor and Grefenstette, Edward},
  journal={Transactions of the Association for Computational Linguistics},
  volume={6},
  pages={317--328},
  year={2018},
  publisher={MIT Press One Rogers Street, Cambridge, MA 02142-1209, USA journals-info~…}
}

@inproceedings{huang2021efficient,
  title={Efficient Attentions for Long Document Summarization},
  author={Huang, Luyang and Cao, Shuyang and Parulian, Nikolaus and Ji, Heng and Wang, Lu},
  booktitle={Proceedings of the 2021 Conference of the North American Chapter of the Association for Computational Linguistics: Human Language Technologies},
  pages={1419--1436},
  year={2021}
}

@inproceedings{shaham2022scrolls,
  title={SCROLLS: Standardized CompaRison Over Long Language Sequences},
  author={Shaham, Uri and Segal, Elad and Ivgi, Maor and Efrat, Avia and Yoran, Ori and Haviv, Adi and Gupta, Ankit and Xiong, Wenhan and Geva, Mor and Berant, Jonathan and others},
  booktitle={Proceedings of the 2022 Conference on Empirical Methods in Natural Language Processing},
  pages={12007--12021},
  year={2022}
}

@inproceedings{shaham2023zeroscrolls,
  title={ZeroSCROLLS: A Zero-Shot Benchmark for Long Text Understanding},
  author={Shaham, Uri and Ivgi, Maor and Efrat, Avia and Berant, Jonathan and Levy, Omer},
  booktitle={Findings of the Association for Computational Linguistics: EMNLP 2023},
  pages={7977--7989},
  year={2023}
}

@article{li2024scbench,
  title={Scbench: A kv cache-centric analysis of long-context methods},
  author={Li, Yucheng and Jiang, Huiqiang and Wu, Qianhui and Luo, Xufang and Ahn, Surin and Zhang, Chengruidong and Abdi, Amir H and Li, Dongsheng and Gao, Jianfeng and Yang, Yuqing and others},
  journal={arXiv preprint arXiv:2412.10319},
  year={2024}
}

@article{kai2025mirbench,
  title={MIR-Bench: Benchmarking LLM’s Long-Context Intelligence via Many-Shot In-Context Inductive Reasoning},
  author={Kai, Yan and Zhan, Ling and Kang, Liu and Yifan, Yang and Ting-Han, Fan and Lingfeng, Shen and Zhengyin, Du and Jiecao, Chen},
  journal={arXiv preprint arXiv:2502.09933},
  year={2025}
}

@article{bai2023longbench,
  title={Longbench: A bilingual, multitask benchmark for long context understanding},
  author={Bai, Yushi and Lv, Xin and Zhang, Jiajie and Lyu, Hongchang and Tang, Jiankai and Huang, Zhidian and Du, Zhengxiao and Liu, Xiao and Zeng, Aohan and Hou, Lei and others},
  journal={arXiv preprint arXiv:2308.14508},
  year={2023}
}

@inproceedings{dong2024bamboo,
  title={BAMBOO: A Comprehensive Benchmark for Evaluating Long Text Modeling Capacities of Large Language Models},
  author={Dong, Zican and Tang, Tianyi and Li, Junyi and Zhao, Wayne Xin and Wen, Ji-Rong},
  booktitle={Proceedings of the 2024 Joint International Conference on Computational Linguistics, Language Resources and Evaluation (LREC-COLING 2024)},
  pages={2086--2099},
  year={2024}
}

@article{liu2024repoqa,
  title={RepoQA: Evaluating Long Context Code Understanding},
  author={Liu, Jiawei and Tian, Jia Le and Daita, Vijay and Wei, Yuxiang and Ding, Yifeng and Wang, Yuhan Katherine and Yang, Jun and Zhang, Lingming},
  journal={arXiv preprint arXiv:2406.06025},
  year={2024}
}

@misc{niah,
    title={Needle In A Haystack - Pressure Testing LLMs},
    author={Greg Kamradt},
    howpublished = {\url{https://github.com/gkamradt/LLMTest_NeedleInAHaystack}},
    year={2023}
}

@article{li2024needlebench,
  title={NeedleBench: Can LLMs Do Retrieval and Reasoning in 1 Million Context Window?},
  author={Li, Mo and Zhang, Songyang and Liu, Yunxin and Chen, Kai},
  journal={arXiv preprint arXiv:2407.11963},
  year={2024}
}

@inproceedings{goldman2024really,
  title={Is It Really Long Context if All You Need Is Retrieval? Towards Genuinely Difficult Long Context NLP},
  author={Goldman, Omer and Jacovi, Alon and Slobodkin, Aviv and Maimon, Aviya and Dagan, Ido and Tsarfaty, Reut},
  booktitle={Proceedings of the 2024 Conference on Empirical Methods in Natural Language Processing},
  pages={16576--16586},
  year={2024}
}

@article{li2024long,
  title={Long-context llms struggle with long in-context learning},
  author={Li, Tianle and Zhang, Ge and Do, Quy Duc and Yue, Xiang and Chen, Wenhu},
  journal={arXiv preprint arXiv:2404.02060},
  year={2024}
}

@article{wang2024novelqa,
  title={Novelqa: A benchmark for long-range novel question answering},
  author={Wang, Cunxiang and Ning, Ruoxi and Pan, Boqi and Wu, Tonghui and Guo, Qipeng and Deng, Cheng and Bao, Guangsheng and Wang, Qian and Zhang, Yue},
  journal={arXiv preprint arXiv:2403.12766},
  year={2024}
}

@article{xu2024detectiveqa,
  title={DetectiveQA: Evaluating Long-Context Reasoning on Detective Novels},
  author={Xu, Zhe and Ye, Jiasheng and Liu, Xiangyang and Sun, Tianxiang and Liu, Xiaoran and Guo, Qipeng and Li, Linlin and Liu, Qun and Huang, Xuanjing and Qiu, Xipeng},
  journal={arXiv preprint arXiv:2409.02465},
  year={2024}
}

@article{ye2025longproc,
  title={LongProc: Benchmarking Long-Context Language Models on Long Procedural Generation},
  author={Ye, Xi and Yin, Fangcong and He, Yinghui and Zhang, Joie and Yen, Howard and Gao, Tianyu and Durrett, Greg and Chen, Danqi},
  journal={arXiv preprint arXiv:2501.05414},
  year={2025}
}

@article{liu2024lost,
  title={Lost in the Middle: How Language Models Use Long Contexts},
  author={Liu, Nelson F and Lin, Kevin and Hewitt, John and Paranjape, Ashwin and Bevilacqua, Michele and Petroni, Fabio and Liang, Percy},
  journal={Transactions of the Association for Computational Linguistics},
  volume={11},
  pages={157--173},
  year={2024}
}

@article{bertsch2024context,
  title={In-context learning with long-context models: An in-depth exploration},
  author={Bertsch, Amanda and Ivgi, Maor and Alon, Uri and Berant, Jonathan and Gormley, Matthew R and Neubig, Graham},
  journal={arXiv preprint arXiv:2405.00200},
  year={2024}
}

@inproceedings{agarwal2024many,
  title={Many-shot In-Context Learning},
  author={Agarwal, Rishabh and Singh, Avi and Zhang, Lei M and Bohnet, Bernd and Rosias, Luis and Chan, Stephanie CY and Zhang, Biao and Faust, Aleksandra and Larochelle, Hugo},
  booktitle={ICML 2024 Workshop on In-Context Learning},
  year={2024}
}

@article{wang2024precision,
  title={When Precision Meets Position: BFloat16 Breaks Down RoPE in Long-Context Training},
  author={Wang, Haonan and Liu, Qian and Du, Chao and Zhu, Tongyao and Du, Cunxiao and Kawaguchi, Kenji and Pang, Tianyu},
  journal={arXiv preprint arXiv:2411.13476},
  year={2024}
}

@inproceedings{hsiehruler,
  title={RULER: What’s the Real Context Size of Your Long-Context Language Models?},
  author={Hsieh, Cheng-Ping and Sun, Simeng and Kriman, Samuel and Acharya, Shantanu and Rekesh, Dima and Jia, Fei and Ginsburg, Boris},
  booktitle={First Conference on Language Modeling},
  url={https://openreview.net/forum?id=kIoBbc76Sy#discussion}
}

@inproceedings{an2024eval,
  title={L-eval: Instituting standardized evaluation for long context language models},
  author={An, Chenxin and Gong, Shansan and Zhong, Ming and Zhao, Xingjian and Li, Mukai and Zhang, Jun and Kong, Lingpeng and Qiu, Xipeng},
  booktitle={Proceedings of the 62nd Annual Meeting of the Association for Computational Linguistics (Volume 1: Long Papers)},
  pages={14388--14411},
  year={2024}
}

@article{bai2024longbench2,
  title={LongBench v2: Towards Deeper Understanding and Reasoning on Realistic Long-context Multitasks}, 
  author={Yushi Bai and Shangqing Tu and Jiajie Zhang and Hao Peng and Xiaozhi Wang and Xin Lv and Shulin Cao and Jiazheng Xu and Lei Hou and Yuxiao Dong and Jie Tang and Juanzi Li},
  journal={arXiv preprint arXiv:2412.15204},
  year={2024},
  url={https://arxiv.org/abs/2412.15204}
}

@article{Karpinska2024OneTA,
  title={One Thousand and One Pairs: A “novel” challenge for long-context language models},
  author={Marzena Karpinska and Katherine Thai and Kyle Lo and Tanya Goyal and Mohit Iyyer},
  journal={ArXiv},
  year={2024},
  volume={abs/2406.16264},
  url={https://arxiv.org/html/2406.16264v1}
}

@inproceedings{yen2025helmet,
      title={HELMET: How to Evaluate Long-Context Language Models Effectively and Thoroughly}, 
      author={Howard Yen and Tianyu Gao and Minmin Hou and Ke Ding and Daniel Fleischer and Peter Izsak and Moshe Wasserblat and Danqi Chen},
      year={2025},
      booktitle={International Conference on Learning Representations (ICLR)},
      url={https://arxiv.org/abs/2410.02694}
}

@inproceedings{NEURIPS2024_babilong,
 author = {Kuratov, Yuri and Bulatov, Aydar and Anokhin, Petr and Rodkin, Ivan and Sorokin, Dmitry and Sorokin, Artyom and Burtsev, Mikhail},
 booktitle = {Advances in Neural Information Processing Systems},
 editor = {A. Globerson and L. Mackey and D. Belgrave and A. Fan and U. Paquet and J. Tomczak and C. Zhang},
 pages = {106519--106554},
 publisher = {Curran Associates, Inc.},
 title = {BABILong: Testing the Limits of LLMs with Long Context Reasoning-in-a-Haystack},
 url = {https://proceedings.neurips.cc/paper_files/paper/2024/file/c0d62e70dbc659cc9bd44cbcf1cb652f-Paper-Datasets_and_Benchmarks_Track.pdf},
 volume = {37},
 year = {2024}
}

@article{Vodrahalli2024MichelangeloLC,
  title={Michelangelo: Long Context Evaluations Beyond Haystacks via Latent Structure Queries},
  author={Kiran Vodrahalli and Santiago Ontannon and Nilesh Tripuraneni and Kelvin Xu and Sanil Jain and Rakesh Shivanna and Jeffrey Hui and Nishanth Dikkala and Mehran Kazemi and Bahare Fatemi and Rohan Anil and Ethan Dyer and Siamak Shakeri and Roopali Vij and Harsh Mehta and Vinay Venkatesh Ramasesh and Quoc Le and Ed Huai-hsin Chi and Yifeng Lu and Orhan Firat and Angeliki Lazaridou and Jean-Baptiste Lespiau and Nithya Attaluri and Kate Olszewska},
  journal={ArXiv},
  year={2024},
  volume={abs/2409.12640},
  url={https://arxiv.org/abs/2409.12640}
}

@inproceedings{modarressi2025nolima,
  title={NoLiMa: Long-Context Evaluation Beyond Literal Matching},
  author={Modarressi, Ali and Deilamsalehy, Hanieh and Dernoncourt, Franck and Bui, Trung and Rossi, Ryan A. and Yoon, Seunghyun and Schütze, Hinrich},
  booktitle={Forty-second International Conference on Machine Learning},
  year={2025},
  url={https://arxiv.org/abs/2502.05167}
}

@misc{openai_mrcr,
  title        = {Introducing GPT-4.1 in the API},
  author       = {{OpenAI}},
  howpublished = {\url{https://openai.com/index/gpt-4-1/}},
  year         = {2025},
}

@article{yang2025controllable,
  title={A Controllable Examination for Long-Context Language Models},
  author={Yang, Yijun and Huang, Zeyu and Zhu, Wenhao and Qiu, Zihan and Yuan, Fei and Pan, Jeff Z and Titov, Ivan},
  journal={arXiv preprint arXiv:2506.02921},
  year={2025}
}

@article{dao2023flashattention,
  title={Flashattention-2: Faster attention with better parallelism and work partitioning},
  author={Dao, Tri},
  journal={arXiv preprint arXiv:2307.08691},
  year={2023}
}

@article{chen2023longlora,
  title={Longlora: Efficient fine-tuning of long-context large language models},
  author={Chen, Yukang and Qian, Shengju and Tang, Haotian and Lai, Xin and Liu, Zhijian and Han, Song and Jia, Jiaya},
  journal={arXiv preprint arXiv:2309.12307},
  year={2023}
}

\clearpage
\beginappendix

\startcontents[app]
\begingroup
  \renewcommand{\contentsname}{Appendix Contents}
  \section*{\contentsname}
  \printcontents[app]{}{1}{}
\endgroup
\newpage

\section{Ethics Statement}

\subsection{Scientific Artifacts}

\textbf{License}: The following are the names of the licenses corresponding to the scientific artifacts we use.
\begin{itemize}
    \item MIT: DeepSeek-V3.2, A-Mem
    \item Apache-2.0: LLMs from the Qwen series (Qwen2.5-7/14B-Instruct-1M, Qwen3-30B-A3B-Instruct-2507, QwenLong), Mem0, MemoryOS
    \item glm-4: GLM-4-9B-Chat-1M
\end{itemize}

\noindent \textbf{Intention of Usage}: Our use of existing artifacts is consistent with their intended use. As for the artifact we create (AgentLong), it is intended solely for research purposes such as the development of Long-Contexts Agents, and shall not be used for any other purposes, including commercial profit-making activities.

\noindent \textbf{Documentation}: For all LLMs we used, please refer to the corresponding model cards or technical reports. For all memory frameworks we used, they are desinged for long-contexts agents and are theoretically applicable to any language.

\subsection{Budget}
In the experiment, the total cost incurred from API calls is about 15632.96 USD.

\section{Dataset Distributions and Statistics}
\label{sec:appendix_distributions}

In this section, we provide a detailed breakdown of the dataset distribution across different settings. We ensure a balanced evaluation by maintaining a consistent number of samples across varying context lengths. The following tables and figures illustrate the distribution of distinct question types for each of the four experimental configurations: \textit{Knowledge-Intensive} vs. \textit{Knowledge-Free}, and \textit{Concise-Response} vs. \textit{Verbose-Response}.

\subsection{Knowledge-Intensive Setting}

In the Knowledge-Intensive setting, tasks involve real-world entities (Pokémon). We visualize the proportion of question types for both formatting strategies in Figure \ref{fig:distribution_ki_concise} and Figure \ref{fig:distribution_ki_verbose}, allowing for a direct comparison between the two structures. Table \ref{tab:dist_ki_concise} and Table \ref{tab:dist_ki_verbose} provide the exact sample counts per context length.

\subsection{Knowledge-Free Setting}

In the Knowledge-Free setting, all entities are masked with abstract tokens to eliminate parametric bias. Figure \ref{fig:distribution_kf_concise} and Figure \ref{fig:distribution_kf_verbose} show the task proportions, while Table \ref{tab:dist_kf_concise} and Table \ref{tab:dist_kf_verbose} detail the sample counts across varying context lengths.

\begin{table*}[h]
    \centering
    \small
    \resizebox{0.95\textwidth}{!}{
    \begin{tabular}{l|cccccccc|c}
    \toprule
    \textbf{Question Type} & \textbf{32K} & \textbf{64K} & \textbf{128K} & \textbf{256K} & \textbf{512K} & \textbf{1M} & \textbf{2M} & \textbf{4M} & \textbf{Total} \\
    \midrule
    Count Frequency (Tool) & 25& 25& 25& 25& 25& 25& 25& 25& 200\\
    Find Duplicates & 25& 25& 25& 25& 25& 25& 25& 25& 200\\
    Find Target Offsets & 25& 25& 25& 25& 25& 25& 25& 25& 200\\
    Count Correctness & 25& 25& 25& 25& 25& 25& 25& 25& 200\\
    Count Frequency (Env) & 16& 12& 18& 17& 23& 21& 16& 17& 140\\
    Find Round with Largest Value& 20& 23& 20& 19& 12& 17& 18& 19& 148\\
    Weighted Summation & 14& 15& 12& 14& 15& 12& 16& 14& 112\\
    Intersection & 50& 50& 50& 50& 50& 50& 50& 50& 400\\
    \midrule
    \textbf{Total} & 200& 200& 200& 200& 200& 200& 200& 200& 1600\\
    \bottomrule
    \end{tabular}}
    \caption{Dataset distribution for Knowledge-Intensive \& Concise-Response. Values represent sample counts.}
    \label{tab:dist_ki_concise}
\end{table*}

\begin{table*}[h]
    \centering
    \small
    \resizebox{0.95\textwidth}{!}{
    \begin{tabular}{l|cccccccc|c}
    \toprule
    \textbf{Question Type} & \textbf{32K} & \textbf{64K} & \textbf{128K} & \textbf{256K} & \textbf{512K} & \textbf{1M} & \textbf{2M} & \textbf{4M} & \textbf{Total} \\
    \midrule
    Count Frequency (Tool) & 25& 25& 25& 25& 25& 25& 25& 25& 200\\
    Find Duplicates & 24& 25& 25& 25& 25& 25& 25& 25& 199\\
    Find Target Offsets & 26& 25& 25& 25& 25& 25& 25& 25& 201\\
    Count Correctness & 25& 25& 25& 25& 25& 25& 25& 25& 200\\
    Count Frequency (Env) & 21& 24& 20& 21& 13& 17& 20& 21& 157\\
    Find Round with Largest Value& 13& 13& 16& 15& 20& 18& 24& 17& 136\\
    Weighted Summation & 16& 13& 14& 14& 17& 15& 6& 12& 107\\
    Intersection & 50& 50& 50& 50& 50& 50& 50& 50& 400\\
    \midrule
    \textbf{Total} & 200& 200& 200& 200& 200& 200& 200& 200& 1600\\
    \bottomrule
    \end{tabular}}
    \caption{Dataset distribution for Knowledge-Intensive \& Verbose-Response. Values represent sample counts.}
    \label{tab:dist_ki_verbose}
\end{table*}

\begin{figure*}[h]
    \centering
    \begin{minipage}{0.48\textwidth}
        \centering
        \includegraphics[width=\linewidth]{figures/en_intersection_pie.png}
        \caption{Distribution for \textit{Knowledge-Intensive} \& \textit{Concise-Response}.}
        \label{fig:distribution_ki_concise}
    \end{minipage}
    \hfill
    \begin{minipage}{0.48\textwidth}
        \centering
        \includegraphics[width=\linewidth]{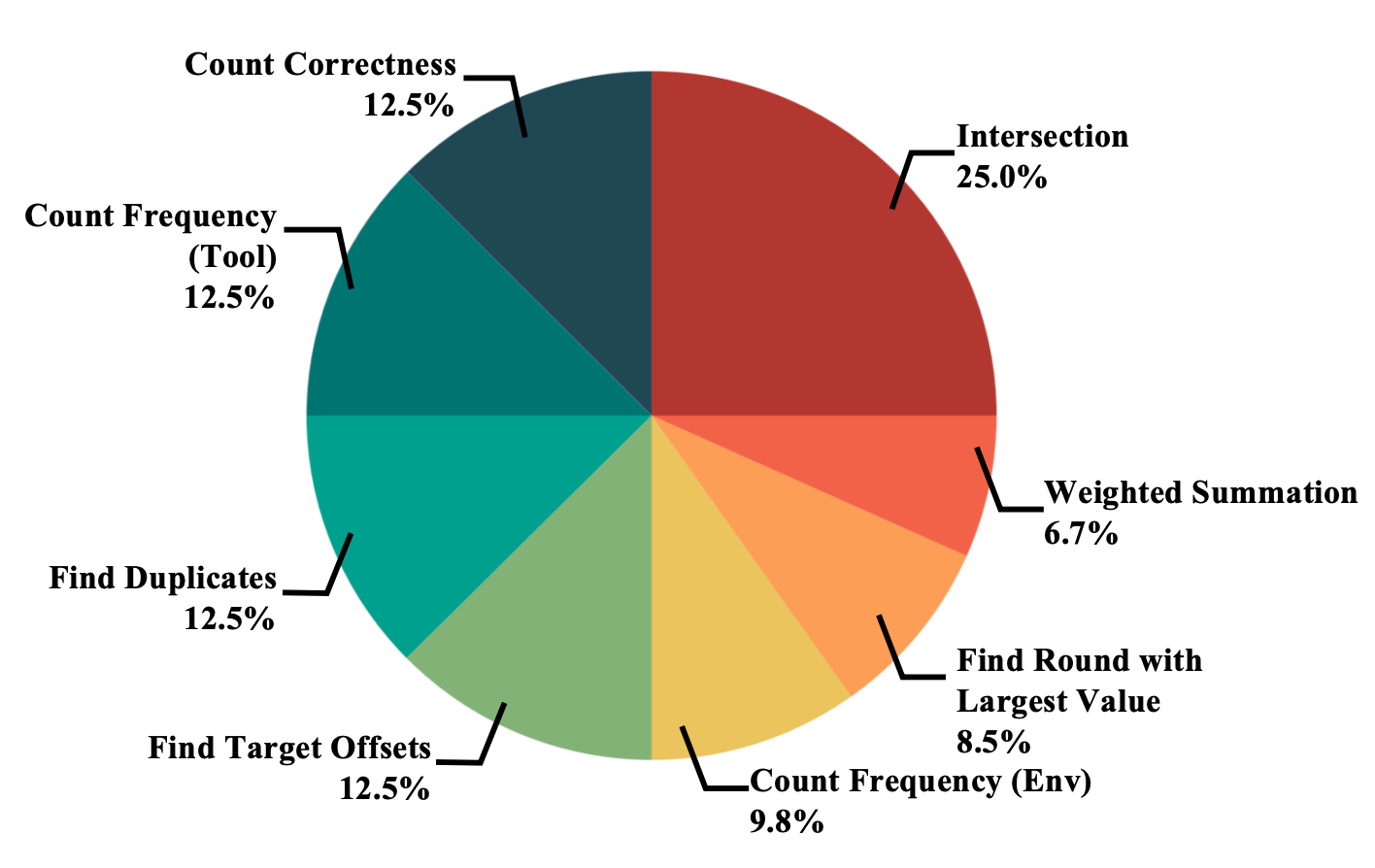}
        \caption{Distribution for \textit{Knowledge-Intensive} \& \textit{Verbose-Response}.}
        \label{fig:distribution_ki_verbose}
    \end{minipage}
\end{figure*}

\begin{figure*}[h]
    \centering
    \begin{minipage}{0.48\textwidth}
        \centering
        \includegraphics[width=\linewidth]{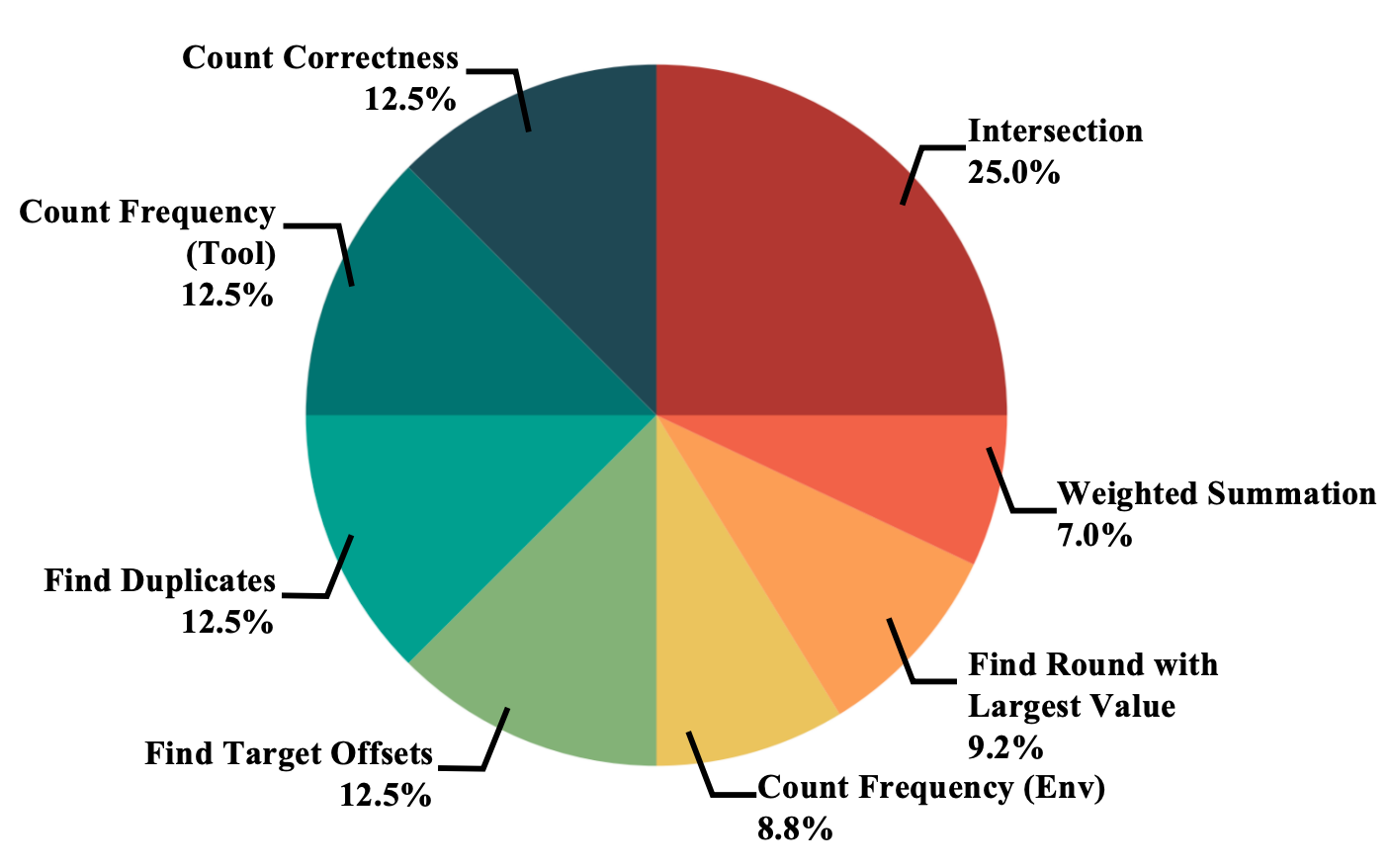}
        \caption{Distribution for \textit{Knowledge-Free} \& \textit{Concise-Response}.}
        \label{fig:distribution_kf_concise}
    \end{minipage}
    \hfill
    \begin{minipage}{0.48\textwidth}
        \centering
        \includegraphics[width=\linewidth]{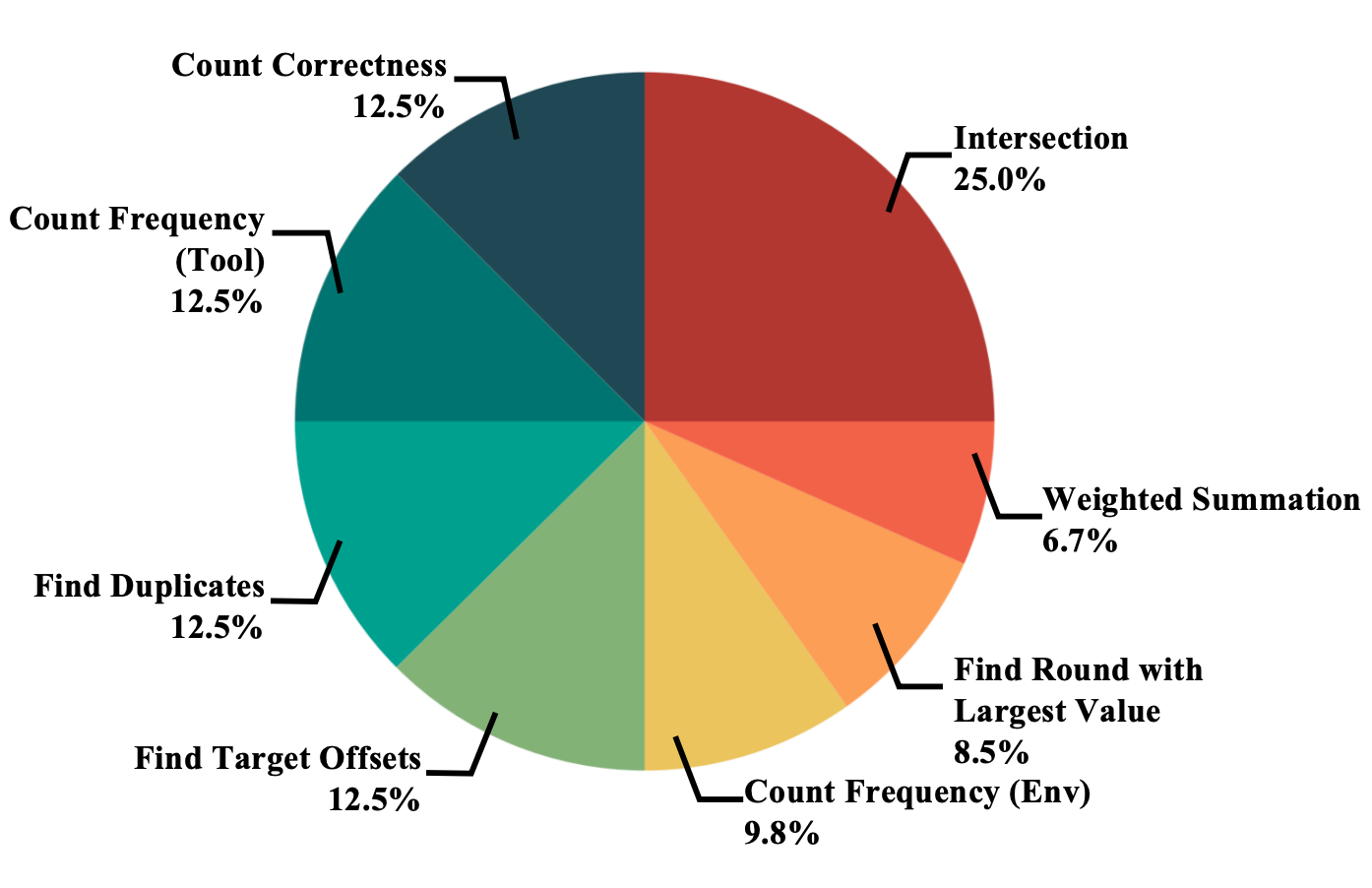}
        \caption{Distribution for \textit{Knowledge-Free} \& \textit{Verbose-Response}.}
        \label{fig:distribution_kf_verbose}
    \end{minipage}
\end{figure*}

\begin{table*}[h]
    \centering
    \small
    \resizebox{0.95\textwidth}{!}{
    \begin{tabular}{l|cccccccc|c}
    \toprule
    \textbf{Question Type} & \textbf{32K} & \textbf{64K} & \textbf{128K} & \textbf{256K} & \textbf{512K} & \textbf{1M} & \textbf{2M} & \textbf{4M} & \textbf{Total} \\
    \midrule
    Count Frequency (Tool) & 25& 25& 25& 25& 25& 25& 25& 25& 200\\
    Find Duplicates & 25& 25& 25& 25& 25& 25& 25& 25& 200\\
    Find Target Offsets & 25& 25& 25& 25& 25& 25& 25& 25& 200\\
    Count Correctness & 25& 25& 25& 25& 25& 25& 25& 25& 200\\
    Count Frequency (Env) & 16& 12& 18& 17& 23& 21& 16& 17& 140\\
    Find Round with Largest Value& 20& 23& 20& 19& 12& 17& 18& 19& 148\\
    Weighted Summation & 14& 15& 12& 14& 15& 12& 16& 14& 112\\
    Intersection & 50& 50& 50& 50& 50& 50& 50& 50& 400\\
    \midrule
    \textbf{Total} & 200& 200& 200& 200& 200& 200& 200& 200& 1600\\
    \bottomrule
    \end{tabular}}
    \caption{Dataset distribution for Knowledge-Free \& Concise-Response.}
    \label{tab:dist_kf_concise}
\end{table*}

\begin{table*}[h]
    \centering
    \small
    \resizebox{0.95\textwidth}{!}{
    \begin{tabular}{l|cccccccc|c}
    \toprule
    \textbf{Question Type} & \textbf{32K} & \textbf{64K} & \textbf{128K} & \textbf{256K} & \textbf{512K} & \textbf{1M} & \textbf{2M} & \textbf{4M} & \textbf{Total} \\
    \midrule
    Count Frequency (Tool) & 25& 25& 25& 25& 25& 25& 25& 25& 200\\
    Find Duplicates & 24& 25& 25& 25& 25& 25& 25& 25& 199\\
    Find Target Offsets & 26& 25& 25& 25& 25& 25& 25& 25& 201\\
    Count Correctness & 25& 25& 25& 25& 25& 25& 25& 25& 200\\
    Count Frequency (Env) & 21& 24& 20& 21& 13& 17& 20& 21& 157\\
    Find Round with Largest Value& 13& 13& 16& 15& 20& 18& 24& 17& 136\\
    Weighted Summation & 16& 13& 14& 14& 17& 15& 6& 12& 107\\
    Intersection & 50& 50& 50& 50& 50& 50& 50& 50& 400\\
    \midrule
    \textbf{Total} & 200& 200& 200& 200& 200& 200& 200& 200& 1600\\
    \bottomrule
    \end{tabular}}
    \caption{Dataset distribution for Knowledge-Free \& Verbose-Response.}
    \label{tab:dist_kf_verbose}
\end{table*}

\clearpage



\section{Trajectory Generation Details}

The data construction pipeline of AgentLongBench is designed to generate scalable, controllable, and logically rigorous interaction logs. The pipeline transforms raw game logic into structured datasets suitable for LLM evaluation. The process consists of four main stages:

\subsection{Base Engine and Simulation Logic}
The core of the generation is a deterministic game engine that manages the ground-truth state. The simulation follows an iterative loop:

\begin{enumerate}
    \item \textbf{Tool Call:} The agent queries specific attributes via the tool API.
    \item \textbf{Tool Result:} The tool returns data based on the formatting strategy (see Section \ref{sec:traj_variants}).
    \item \textbf{Model Guess:} The simulated agent proposes a candidate item.
    \item \textbf{Engine Feedback:} The environment compares the guess with the hidden target and provides differential feedback.
\end{enumerate}

\subsection{Trajectory Variants: Concise vs. Verbose}
\label{sec:traj_variants}
We generate two distinct interaction histories from the same underlying game engine to isolate the effects of memory fragmentation versus information density. Figures \ref{fig:traj_concise} and \ref{fig:traj_verbose} provide side-by-side examples of the raw JSON logs used in these configurations.

\noindent \textbf{Concise-Response.} As illustrated in Figure \ref{fig:traj_concise}, the tool in this setting functions as a logic filter, returning only the intersection of items that satisfy all queried attributes. By abstracting away per-section candidate lists, this format maintains a low token density per turn. Consequently, interaction histories can extend to hundreds of rounds, shifting the cognitive burden to long-term state tracking and testing the model's resilience against memory fragmentation.

\noindent \textbf{Verbose-Response.} Conversely, the Verbose format (Figure \ref{fig:traj_verbose}) mimics raw database outputs where the tool returns independent, unfiltered candidate lists for each queried condition. This approach preserves all raw evidence but floods the context window with high-density structured noise. Although this results in fewer interaction rounds for a given token budget, it compels the agent to perform logical intersections internally, thereby strictly testing its capacity to handle information overload within a single reasoning step.

\subsection{Behavioral Control Parameters}
To ensure the generated trajectories mimic realistic and imperfect agent behaviors rather than optimal searches, we introduce several control parameters into the simulator:

\begin{description}
    \item[\texttt{history\_window}:] Defines a rolling window size representing the agent's active working memory. Conditions outside this window may be dropped.
    \item[\texttt{forget\_history\_prob}:] The probability that older conditions outside the current batch are "forgotten" by the simulated agent. Higher values simulate an agent with poorer long-term retention, necessitating redundant queries.
    \item[\texttt{mask\_prob} \& \texttt{max\_mask\_sections}:] Parameters that randomly hide specific sections or conditions in the tool query, simulating partial attention or incomplete information gathering.
    \item[\texttt{epsilon} (Exploration Rate):] The probability of "relaxing" constraints. Instead of strictly querying the intersection, the agent may explore broader attributes. This prevents the trajectory from converging too quickly, allowing for longer contexts.
\end{description}

\subsection{Post-Processing and Task Construction}

\paragraph{Token-Length Truncation}
Raw histories are essentially infinite. We truncate and bucket these trajectories into fixed context lengths (32K, 64K, ..., 4M). Crucially, this process preserves \textbf{whole rounds only} to ensure logical integrity. 

\paragraph{Final Guess Specialization}
For the \textit{Intersection} task in concise-response scene, simple truncation is insufficient because the final answer must be logically deducible from the remaining context. We employ a strict filtering process for this task:
\begin{itemize}
    \item Ensure the intersection of all tool responses in the history yields exactly the unique target item.
    \item Verify that every round's intersection list is sufficiently large to maintain task difficulty before the final convergence.
\end{itemize}

\paragraph{QA Dataset Generation}
The final benchmark samples are created by selecting specific target rounds from the processed histories. For each sample, we assemble the full `messages` list (System, User, Assistant, Tool) up to the target point and attach the corresponding question (e.g., "Find Duplicates") and the deterministic ground-truth answer.

\subsection{Knowledge-Free Adaptation}
The \textit{Knowledge-Free} setting is derived directly from the pipeline above. We apply a symbolic mapping layer where all semantic entities (e.g., Pokémon names, types, ability names) are replaced with abstract tokens (e.g., \texttt{Item\_29A}, \texttt{Attr\_B}). This ensures that the structural complexity and logical dependencies of the dataset remain identical to the \textit{Knowledge-Intensive} version, while strictly isolating the model's reasoning capabilities from its parametric knowledge.

\section{Detailed Tasks Taxonomy}
\label{sec:appendix_tasks}

We design three categories of tasks comprising eight distinct question types. These tasks allow us to isolate specific cognitive failures in long-context agents.

\subsection{QA in Tool Response}
This category evaluates the agent's ability to recall specific details from tool execution outputs, testing robustness in parsing machine-generated data.
\begin{itemize}
    \item \textbf{Count Frequency:} Count the frequency of a specific item appearing in the tool return values for a specific round. Unlike simple retrieval, the ``needle'' here is embedded in structured tool noise.
    \item \textbf{Find Duplicates:} Determine whether a specific item appears in the tool return values of \textbf{both} Round $i$ and Round $j$. This evaluates retrieval across temporal distances.
    \item \textbf{Find Target Offsets:} Identify the two items immediately following the first occurrence of a specific item in the tool return list for Round $i$. This tests positional matching ability.
\end{itemize}

\subsection{QA in Environment Response}
This category focuses on the interaction history with the Host, testing the agent's ability to track the \textit{State} and ``Yes/No'' constraints.
\begin{itemize}
    \item \textbf{Count Correctness:} Determine how many attribute sections were guessed correctly in a specific round based on feedback.
    \item \textbf{Count Frequency:} Count how many times a specific attribute value has appeared across the feedback of \textbf{all} rounds.
    \item \textbf{Find Round with Largest Value:} Identify which round's feedback contained the highest specific numeric attribute value.
    \item \textbf{Weighted Summation:} Calculate the absolute difference between the weighted scores of Round $i$ and Round $j$. The score is calculated based on a weighted scheme assigning points to different attribute categories (e.g., Type: 6, Ability: 5, Base Stats: 4, etc.), testing both retrieval and computational reasoning.
\end{itemize}

\subsection{Final Guess}
This predictive category requires the agent to generate the final answer based on the global understanding of the context.
\begin{itemize}
    \item \textbf{Intersection:} The agent must compute the intersection of candidate items based on the provided tool responses. 
    \begin{itemize}
        \item In the \textbf{Concise-Response} format, since the tool returns pre-filtered results, this effectively requires intersecting constraints across the entire history to deduce the final target. 
        \item In the \textbf{Verbose-Response} format, this requires identifying the intersection of attribute lists within a specific round's tool response.
    \end{itemize}
\end{itemize}
This task evaluates the agent's ability to perform logical set operations and synthesize information, determining whether it understands the evolving state rather than merely retrieving facts.

\section{Baselines and Implementation Details}
\label{sec:baselines}

Our experimental evaluation covers a comprehensive spectrum of proprietary frontiers, open-weight models, and specialized memory architectures. All inference tasks were executed using the VLLM framework to ensure high-throughput processing. We maintained a consistent sampling temperature of 0.7 across all runs to balance generation diversity with instruction adherence.

\paragraph{Proprietary Models}
We accessed proprietary systems via their official APIs. To ensure fair comparison, we utilized the standard chat templates and default system prompts recommended by the respective vendors, without specific prompt engineering for context compression.

\paragraph{Open-Source Models}
For open-weight baselines, we deployed the official Instruct versions of the Qwen series (Qwen2.5, Qwen3) and GLM-4 on a cluster of NVIDIA H200 GPUs. In cases where models support multiple system prompt configurations, we unified the evaluation under a standardized prompt to isolate the model's intrinsic context processing capabilities.

\paragraph{Memory and RAG Frameworks}
For Retrieval-Augmented Generation (RAG) and dedicated agentic memory systems (A-Mem, Mem0, MemoryOS), we adhered strictly to their public default configurations to reflect standard usage patterns. We standardized the retrieval hyperparameter across all memory-augmented baselines, setting the top-$k$ retrieved chunks to 5 ($k=5$). This setting provides a consistent context window for the backbone model (Qwen3-30B-A3B) to synthesize the final response.

\begin{figure*}[t]
    \centering
    \textbf{Trajectory Variant A: Concise-Response (Intersection Only)}
    \begin{lstlisting}[language=json, captionpos=b]
{
  "messages": [
    {
      "role": "system",
      "content": "You are playing a guess-the-Pokemon game... (*@\textit{\color{gray}[System prompt truncated]}@*)..."
    },
    {
      "role": "user",
      "content": "Round 1: Guess Kakuna (#0014)\nSections:\n - Type: Bug (wrong); Poison (wrong)... (*@\textit{\color{gray}[Attributes Omitted]}@*)... \nResult: wrong\nRemaining rounds: 2009"
    },
    {
      "role": "assistant",
      "content": "<think>Thinking and calling query_pokemon.</think>",
      "tool_calls": [
        {
          "id": "call_3",
          "type": "function",
          "function": {
            "name": "query_pokemon",
            "arguments": "{\"conditions\": [{\"type\": \"value\", \"section\": \"Abilities\", \"values\": [\"Shed Skin\"], \"exclude\": true}, {\"type\": \"value\", \"section\": \"Type\", \"values\": [\"Bug\", \"Poison\"], \"exclude\": true}]}"
          }
        }
      ]
    },
    {
      "role": "tool",
      "tool_call_id": "call_3",
      "name": "query_pokemon",
      "content": "{\"intersection\": [\"Abomasnow\", \"Abra\", ..., \"Zygarde\"]}"
      (*@\textit{\color{gray}// Concise: The tool executes the logic and returns ONLY the intersection list.}@*)
    },
    {
      "role": "assistant",
      "content": "<answer>Thwackey</answer>"
    },
    (*@\textit{\color{gray}... [Rounds 2-10 omitted for brevity] ...}@*)
    {
      "role": "user",
      "content": "Round 11: Guess Kirlia (#0281)\nSections:\n - Type: Psychic (wrong); Fairy (wrong)\n - Base Stats: 278 (wrong, too low)...\nResult: wrong\nRemaining rounds: 1999"
    },
    {
      "role": "assistant",
      "content": "<think>Filtering based on new constraints...</think>",
      "tool_calls": [
        {
          "id": "call_43",
          "type": "function",
          "function": {
            "name": "query_pokemon",
            "arguments": "{\"conditions\": [{\"type\": \"value\", \"section\": \"Type\", \"values\": [\"Psychic\", \"Fairy\"], \"exclude\": true}, {\"type\": \"numeric\", \"section\": \"Base Stats\", \"comparator\": \">\", \"threshold\": 278}]}"
          }
        }
      ]
    },
    {
      "role": "tool",
      "tool_call_id": "call_43",
      "name": "query_pokemon",
      "content": "{\"intersection\": [\"Abomasnow\", \"Absol\", ..., \"Zweilous\"]}"
    },
    {
      "role": "assistant",
      "content": "<answer>Darkrai</answer>"
    }
  ]
}
    \end{lstlisting}
    \vspace{-0.5em}
    \caption{Concise-Response Trajectory Example.}
    \label{fig:traj_concise}
\end{figure*}

\begin{figure*}[t]
    \centering
    \textbf{Trajectory Variant B: Verbose-Response (Full Candidate Lists)}
    \begin{lstlisting}[language=json, captionpos=b]
{
  "messages": [
    {
      "role": "system",
      "content": "You are playing a guess-the-Pokemon game... (*@\textit{\color{gray}[System prompt truncated]}@*)..."
    },
    {
      "role": "assistant",
      "content": "<answer>Seedot</answer>"
    },
    {
      "role": "user",
      "content": "Round 1: Guess Seedot (#0273)\nSections:\n - Type: Grass (correct)\n ... (*@\textit{\color{gray}[Attributes Omitted]}@*)...\nResult: wrong\nRemaining rounds: 259"
    },
    {
      "role": "assistant",
      "content": "<think>Thinking and calling query_pokemon.</think>",
      "tool_calls": [
        {
          "id": "call_3",
          "type": "function",
          "function": {
            "name": "query_pokemon",
            "arguments": "{\"conditions\": [{\"type\": \"value\", \"section\": \"Type\", \"values\": [\"Grass\"]}, {\"type\": \"numeric\", \"section\": \"Base Stats\", \"comparator\": \">\", \"threshold\": 220}, {\"type\": \"numeric\", \"section\": \"Generation\", \"comparator\": \"==\", \"threshold\": 3}, ...]}"
          }
        }
      ]
    },
    {
      "role": "tool",
      "tool_call_id": "call_3",
      "name": "query_pokemon",
      "content": {
        "per_section": [
          {
            "section": "Type",
            "conditions": [{"type": "value", "values": ["Grass"]}],
            "candidates": [
              "Abomasnow", "Amoonguss", "Appletun", "Applin", "Arboliva",(*@\textit{\color{gray}... [138 items omitted] ...}@*),"Zarude"
            ]
          },
          {
            "section": "Base Stats",
            "conditions": [{"type": "numeric", "comparator": ">", "threshold": 220}],
            "candidates": [
              "Abomasnow", "Abra", "Absol", "Accelgor", "Aegislash",(*@\textit{\color{gray}... [850 items omitted] ...}@*),"Zygarde"
            ]
          },
          {
            "section": "Generation",
            "conditions": [{"type": "numeric", "comparator": "==", "threshold": 3}],
            "candidates": [ (*@\textit{\color{gray}... [135 items omitted] ...}@*) ]
          },
          {
            "section": "Abilities",
            "candidates": [ (*@\textit{\color{gray}... [Huge list omitted] ...}@*) ]
          }
        ]
      }
      (*@\textit{\color{gray}// Verbose: The tool returns SEPARATE lists for each condition. The agent must mentally intersect these hundreds of items to find the target.}@*)
    },
    {
      "role": "assistant",
      "content": "<answer>Lileep</answer>"
    },
    (*@\textit{\color{gray}... [Rounds 2-3 omitted] ...}@*)
    {
      "role": "user",
      "content": "Round 4: Guess Palossand (#0770)\nSections:\n - Type: Ghost (wrong); Ground (wrong)\n - Base Stats: 480 (wrong, too high)...\nResult: wrong\nRemaining rounds: 256"
    }
  ]
}
    \end{lstlisting}
    \vspace{-0.5em}
    \caption{Verbose-Response Trajectory Example.}
    \label{fig:traj_verbose}
\end{figure*}

\end{document}